\documentclass{article}

 \PassOptionsToPackage{numbers, sort&compress}{natbib}
\usepackage[final]{neurips_2021}

\usepackage[utf8]{inputenc} % allow utf-8 input
\usepackage[T1]{fontenc}    % use 8-bit T1 fonts
\usepackage{xr-hyper}
\usepackage{hyperref}       % hyperlinks
\usepackage{url}            % simple URL typesetting
\usepackage{booktabs}       % professional-quality tables
\usepackage{amsfonts}       % blackboard math symbols
\usepackage{nicefrac}       % compact symbols for 1/2, etc.
\usepackage{microtype}      % microtypography
\usepackage[dvipsnames]{xcolor}         % colors
\usepackage{natbib}
\usepackage{amsmath,amssymb}
\usepackage{graphicx}
\usepackage{caption}
\usepackage{enumitem}
\usepackage{subcaption}
\usepackage{diagbox}
\usepackage{soul}
\usepackage{svg}
\usepackage{wrapfig}
\usepackage{floatrow}
\usepackage[title]{appendix}

\newfloatcommand{capbtabbox}{table}[][\FBwidth]
\makeatletter
\newcommand*{\addFileDependency}[1]{% argument=file name and extension
  \typeout{(#1)}
  \@addtofilelist{#1}
  \IfFileExists{#1}{}{\typeout{No file #1.}}
}
\makeatother

\usepackage{pdfpages}
\title{Grounding inductive biases in natural images: invariance stems from variations in data}
\author{Diane Bouchacourt\thanks{equal contribution, a coin was flipped},~~Mark Ibrahim\footnotemark[1], ~Ari S. Morcos \\
Facebook AI Research\\
\texttt{\{dianeb,marksibrahim,arimorcos\}@fb.com} \\
}

\begin{document}

\maketitle

\begin{abstract}
To perform well on unseen and potentially out-of-distribution samples, it is desirable for machine learning models to have a predictable response with respect to transformations affecting the factors of variation of the input. Here, we study the relative importance of several types of inductive biases towards such predictable behavior: the choice of data, their augmentations, and model architectures. Invariance is commonly achieved through hand-engineered data augmentation, but do standard data augmentations address transformations that explain variations in real data? While prior work has focused on synthetic data, we attempt here to characterize the factors of variation in a real dataset, ImageNet, and study the invariance of both standard residual networks and the recently proposed vision transformer with respect to changes in these factors. We show standard augmentation relies on a precise combination of translation and scale, with translation recapturing most of the performance improvement---despite the (approximate) translation invariance built in to convolutional architectures, such as residual networks. In fact, we found that scale and translation invariance was similar across residual networks and vision transformer models despite their markedly different \textit{architectural} inductive biases. We show the training data itself is the main source of invariance, and that data augmentation only further increases the learned invariances. Notably, the invariances learned during training align with the ImageNet factors of variation we found. Finally, we find that the main factors of variation in ImageNet mostly relate to appearance and are specific to each class.
%  Characterizing the factors of variation present in realistic data is challenging and this question has mostly been addressed in synthetic (and often simplistic) settings with full access to the generative model of the data. 
%  enhances a model's natural invariance for transformations that account for ImageNet variations, while it has the opposite effect for changes that do not affect factors of variation. 
\end{abstract}

\section{Introduction}
\label{sec:intro}
A dataset can be described in terms of natural factors of variation of the data: for example, images of objects can present those objects with different poses, illuminations, colors, etc. Prediction consistency of models with respect to changes in these factors is a desirable property for out-of-domain generalization \citep{bengio_representation_2013,lenc18understanding}. However, state-of-the-art Convolutional Neural Networks (CNNs) struggle when presented with ``unusual" examples, e.g. a bus upside down \citep{alcorn2019strike}. Indeed, CNNs lack robustness not only to changes in pose, but even to simple geometric transformations such as small translations and rotations \citep{Engstrom19, Azulay19}.

\begin{figure}[t]
	\centering
    \begin{subfigure}{.5\textwidth}
    	\includegraphics[width=\textwidth]{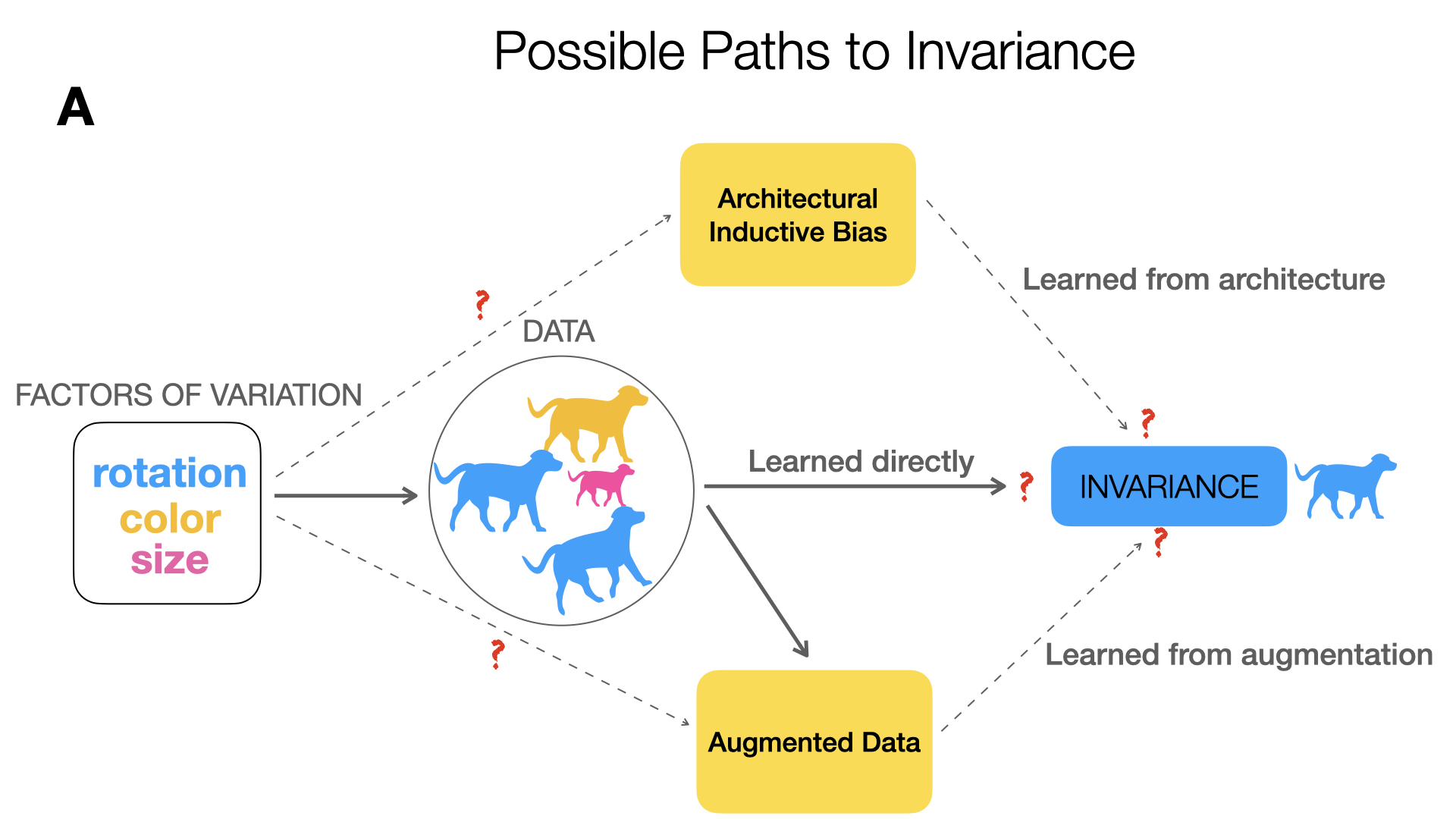}
    \end{subfigure}%
    \begin{subfigure}{.5\textwidth}
    	\includegraphics[width=\textwidth]{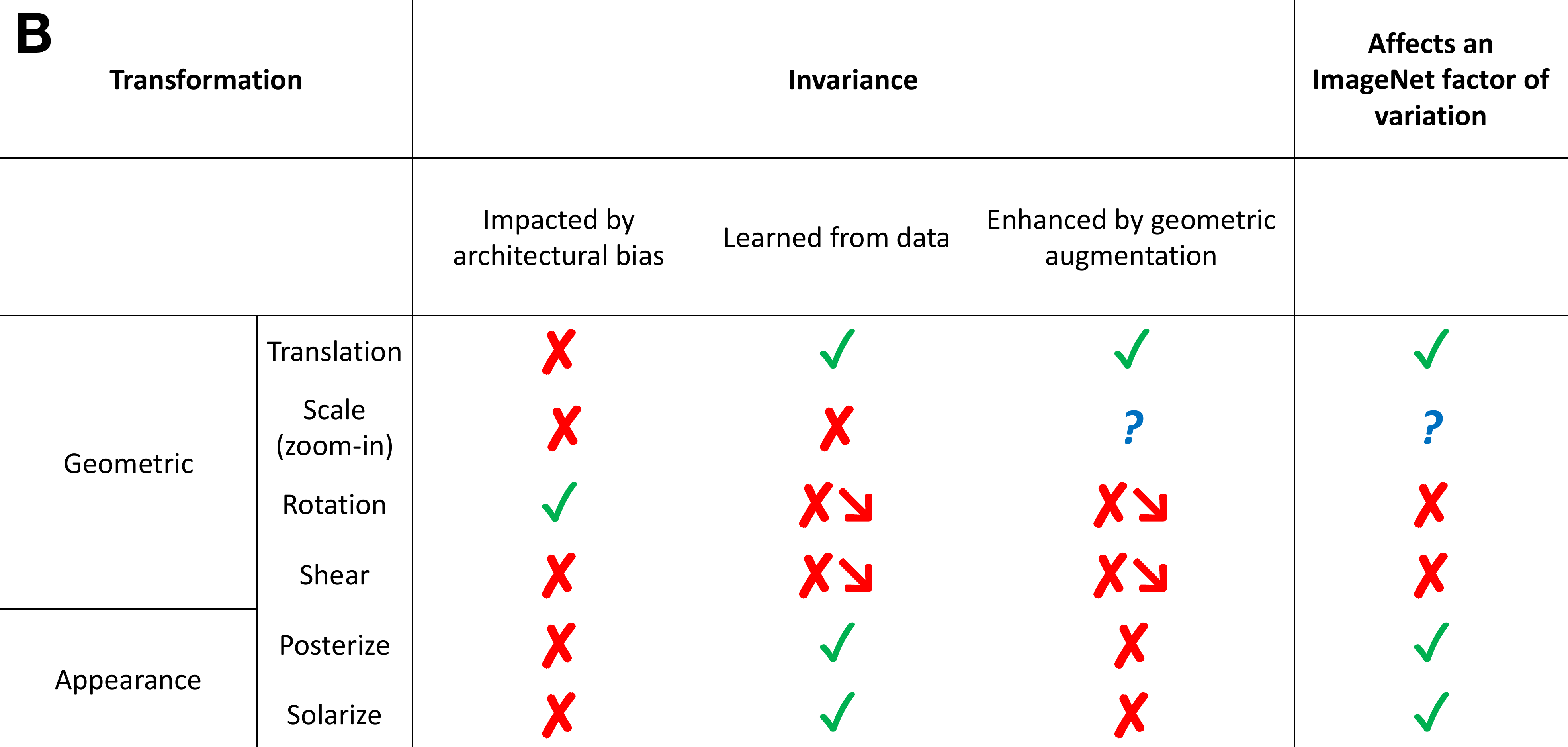}
    \end{subfigure}
    \caption{\textbf{Grounding invariance in factors of variation}. In Fig. A, we illustrate possible paths to learning invariance: architectural inductive bias, directly from data, and through data augmentation. In Fig. B, we summarize our findings about invariance. A red arrow means invariance decreases, a question mark means results do not allow for a clear answer.}
    \label{fig:introfig}
\end{figure}

Invariance to factors of variation can be learned directly from data, built-in via architectural inductive biases, or encouraged via data augmentation (Fig. \ref{fig:introfig}A). Our goal with this work is to explore the relative impact of these three factors on the trained model's learned invariances and performance. As of today, data augmentation is the predominant method for encouraging invariance to a set of transformations. Yet, even with data augmentation models fail to generalize to held out objects and to learn invariance. For example, \citep{Engstrom19} found that augmenting with rotation and translation does not lead to invariance to the very same transformations during testing. A complementary research direction ensures a model responds predictably to transformations, using group equivariance theory (see \citet{pmlr-v48-cohenc16, cohen_general_2020} among others). 
% Later network layers can disregard the predictable responses to induce invariance. 
Provably invariant models have limited large-scale applications as they require a priori knowledge of the underlying factors of variation \citep{pmlr-v48-cohenc16, finzi_generalizing_2020}. Recent work tackles automatic discovery of symmetries in data \citep{benton2020learning,zhou_meta-learning_2020,dehmamy2021lie, Hashimoto2017}. However, these methods have mostly been applied to synthetic or artificially augmented datasets which are not directly transferable to real data settings, and can even hurt performance when transferred \citep{dehmamy2021lie}. Thus, we aim to identify the factors of variation of a real image dataset and to understand whether such equi/invariant models would be a relevant choice.

One way to characterize the consistency of a model's response is by measuring its equivariance or invariance to changes of the input. A model, $f$, is equivariant to a transformation $T_\theta$ of an input $x$ if the model's output transforms in a corresponding manner via an output transformation $T_\theta'$, i.e. $T'_\theta(f(x)) = f(T_\theta(x))$ for any $x$. Invariance is a special type of equivariance, where the model's output is the same for all transformations, i.e. $T_{\theta}'=\text{I}$. Throughout this work, we will use for $f$ the penultimate layer of a Resnet18 \citep{He2015} or a vision transformer \citep{dosovitskiy2020image} trained on ImageNet. To understand whether CNNs and other state-of-the-art models are invariant to changes in the data factors of variation, one needs explicit annotations about such factors. While this is trivial for synthetic datasets with known factors, identifying the factors of variations in real datasets is a complex task. As such, prior work turned to synthetic settings to show that knowing the underlying factors improves generalization \citep{ChengRotDCF, locatello_weakly-supervised_2020}. 

Here, we take a first step towards understanding the links between data augmentations and factors of variation in natural images, in the context of image classification. We do so by carefully studying the role of data augmentation, architectural inductive biases, and the data itself in encouraging invariance to these factors. We primarily focus on ResNet18 trained ImageNet as a benchmark for large-scale vision models \citep{deng2009imagenet, He2015}, which we also compare to the recently proposed vision transformer (ViT) \citep{dosovitskiy2020image}. While previous works study the invariance properties of neural networks to a set of transformations \citep{lenc18understanding, Myburgh2021, Zhang2019, Kauderer2018, Engstrom19},
% \DB{Remove citations and refer to related work section?}, 
we ground invariances in ImageNet factors of variations. We make the following contributions (summarized in Fig. \ref{fig:introfig}):
\begin{itemize}[topsep=0pt]
    \item \textit{What transformations do standard data augmentation correspond to?} In Sec. \ref{sec:dataaugBig}, we demonstrate that the success of the popular random resized crop (RRC) augmentation amounts to a precise combination of translation and scaling. To tease out the relative role of these factors, we decomposed RRC into separate augmentations. While neither augmentation alone was sufficient to fully replace RRC, we observed that despite the approximate translation invariance built into CNNs, translation alone is sufficient to improve performance close to RRC, whereas the contribution of scale was comparatively minor.
    % Moreover, there is a trade-off between variability and the magnitude of the augmentations used.
    \item \textit{What types of invariance are present in ImageNet trained models? Do these invariances derive from the data augmentation, the architectural bias or the data itself (Figure \ref{fig:introfig}A)?} In Sec. \ref{sec:invariance}, we demonstrate that when invariance is present, it is primarily learned from data independent of the augmentation strategy used with the notable exception of translation invariance which is enhanced by standard data augmentation. We also found that architectural bias has a minimal impact on invariance to the majority of transformations.
    \item \textit{Which transformations account for intra-class variations in ImageNet? How do they relate to the models' invariance properties discovered in Sec. \ref{sec:invariance}?} In Sec. \ref{sec:simsearch2}, we show that appearance transformations, often absent from standard augmentations, account for intra-class changes for factors of variation in ImageNet. We found training enhances a model's natural invariance to transformations that account for ImageNet variations (including appearance transformations), and decreases invariance for transformations that do not seem to affect factors of variation. We also found factors of variation are unique per class, despite common data augmentations applying the same transformations across all classes.
\end{itemize}

Our results demonstrate that the relationship among architectural inductive biases, the data itself, the augmentations used, and invariance is often more complicated than it may first appear, even when the relationship appears intuitive (such as for convolution and translation invariance). Furthermore, invariance generally stems from the data itself, and aligns with the data factors of variations. By understanding both which invariances are desirable and how to best encourage these invariances, we hope to guide future research into building more robust and generalizable models for large scale, realistic applications. Code to reproduce our results is in supplementary material. 

\section{Decomposing the Random Resized Crop Augmentation}
\label{sec:dataaugBig}

Data augmentation improves performance and generalization by transforming inputs to increase the amount of training data and its variations \citep{Niyogi98incorporatingprior}. Transformations typically considered include taking a crop of random size and location (random resized crop), horizontal flipping, and color jittering \citep{He2015, Touvron2019, Simonyan15}. Here, we focus on random resized crop (denoted \texttt{RandomResizedCrop}; RRC) that is commonly used for training ResNets \footnote{We follow the procedure of \texttt{RandomResizedCrop} as implemented in the PyTorch library \citep{PyTorch} \texttt{https://pytorch.org/vision/stable/\_modules/torchvision/transforms/transforms.html}.}. 
% Here, we focus on geometric transformations commonly used for training ResNets. Specifically, we study random resized crop (denoted \texttt{RandomResizedCrop}; RRC)\footnote{We follow the procedure of \texttt{RandomResizedCrop} as implemented in the PyTorch library \citep{PyTorch} \texttt{https://pytorch.org/vision/stable/\_modules/torchvision/transforms/transforms.html}.}. 

For an image of width $W$ and height $H$, \texttt{RandomResizedCrop} (1) samples a scale factor $s$ from a uniform distribution, $s \sim U(s_{-}, s_{+})$ and an aspect ratio $r\sim U(\ln r_{-},\ln r_{+})$ (2) takes a square crop of size $\sqrt{sHWr} \times \sqrt{sHW/r}$ in any part of the image (3) resizes the crop, typically to $224\times224$ for ImageNet. Thus, the area of an object selected by the crop is randomly scaled proportional to 1/$s$, which encourages the model to be scale invariant. The crop is also taken in any location of the image within its boundaries, which is equivalent to applying a translation whose parameters depend on the percentage of the area chosen for the crop. However, the way translation and scale interact remains obscure. In this section, we contrast the role of translation and scale in RRC and analyze the impact of the parameters used to determine these augmentations.
% corresponding to a $\%$ of the image area
\subsection{The gain of \texttt{RandomResizedCrop} is largely driven by translation rather than scale}
\label{sec:dataaug}
To study the role of both the scaling and translation steps, we separate \texttt{RandomResizedCrop} into two component data augmentations: 
\begin{itemize}[noitemsep,topsep=0pt]
    \item \texttt{RandomSizeCenterCrop} takes a crop of random size, always at the center of the image. The distribution for scale and aspect ratio are the same as those used in \texttt{RandomResizedCrop}. This augmentation impacts scale, but not translation.
    \item \texttt{FixedSizeRandomCrop} takes a crop of fixed size ($224\times 224$) at any location of the image (the image is first resized to $256$ on the shorter dimension). This augmentation impacts translation, but not scale. 
\end{itemize}
\begin{table}[t!]
\setlength{\tabcolsep}{3pt} % Default value: 6pt
\begin{subtable}[t]{0.65\textwidth}
	\centering
    {\begin{tabular}[t]{l|c}
    Method & {Top-1 $\pm$ SEM}\\
    \hline
    \texttt{RandomResizedCrop} (RRC) & $\mathbf{70.05\pm 0.1}$\\
    \texttt{RandomSizeCenterCrop}  & $67.84\pm 0.1$\\
    \texttt{FixedSizeRandomCrop} & $67.93\pm 0.0$\\
    \texttt{T.(30\%)} & $69.14\pm 0.0$ \\
    \texttt{T.(30\%) + RandomSizeCenterCrop} &$69.30\pm 0.0$\\
    \texttt{T.(30\%) + RandomSizeCenterCrop} w/o a.r. & $69.20\pm 0.1$\\
    \texttt{FixedSizeCenterCrop} (no augmentation) & $63.49\pm 0.1$\\
    \end{tabular}}
    \caption{Using different training augmentations used, w/o a.r. stands for without aspect ratio change, T. stands for Translation.}
    \label{tab:augmentfull}
\end{subtable}
\hfill
\begin{subtable}[t]{0.32\textwidth}
    \centering
    {\begin{tabular}[t]{c|c}
      $\beta$ & \multicolumn{1}{c}{Top-1 $\pm$ SEM}\\
      \hline
        0.1 & $69.32\pm 0.0$\\
        0.5 & $\mathbf{70.17\pm 0.1}$\\
        1 ($\sim$ RRC) & $\mathbf{70.11\pm 0.0}$\\
        2 & $69.52\pm 0.1$\\
        3 & $68.79\pm 0.0$\\
        10 & $63.74\pm 0.1$\\
    \end{tabular}}
    \caption{Varying the shape of the distribution over $s$. RRC stands for \texttt{RandomResizedCrop}.}
    \label{tab:CCTcomp}
\end{subtable} 
\caption{ImageNet validation Top-1 accuracy $\pm$ SEM (standard error of the mean).}
\end{table}
As with RRC, both of these transformations can remove information from the image (for example, translation can shift a portion of the image out of the frame whereas zooming in will remove the edges of the image), but neither augmentation can fully reproduce the effect of RRC by itself.

We train ResNet18 on ImageNet and report results on the validation set as in commonly done (e.g. in \citet{Touvron2019}) since the labelled test set is not publicly available. \texttt{FixedSizeCenterCrop} corresponds to what is usually done for augmenting validation/test images, i.e. resize the image to $256$ on the shorter dimension and take a center crop of size $224$. We apply \texttt{FixedSizeCenterCrop} during all evaluation steps and refer to it as ``no augmentation". Training details are in Appendix \ref{sec:trainingdetails}.

\textbf{RCC combines scale and translation in a precise manner.} Table \ref{tab:augmentfull} shows that \texttt{RandomResizedCrop} performs best, with $70.05\%$ Top-1 accuracy. Augmenting by taking a crop of random size (\texttt{RandomSizeCenterCrop}), a proxy for scale-invariance of the center object, performs on par with \texttt{FixedSizeRandomCrop}, a proxy to invariance to translation, and both bring a substantial improvement compared to no augmentation (~67.9\% vs. ~63.5\%). However, neither is sufficient to fully recapture the performance of RRC. To further test that RRC amounts to translation and scale, we augment the image by translating it by at most 30\% 
% (random sampling from $-30\%$ to $30\%$) 
in width and height 
% without periodic boundaries,
followed by taking a center crop of random size (denoted \texttt{T.(30\%) + RandomSizeCenterCrop}). This method improves Top-1 accuracy over \texttt{RandomSizeCenterCrop} and \texttt{FixedSizeRandomCrop}, and almost match RRC but with a gap of $0.75\%$, showing that scaling and translating interact in a precise manner in RRC that is difficult to reproduce with both transformations applied iteratively.

\textbf{RCC's performance is driven by translation.} \texttt{FixedSizeRandomCrop} impacts translation, but its behavior is contrived, for example an image in the top left corner will never be in the bottom left corner of a crop. Thus to further disambiguate the role of translation versus scale, we also experiment using \texttt{T.(30\%)} only: we resize the image to $256$ on the shorter dimension, apply the random translation of at most $30\%$ and take a center crop of size $224\times224$. 
% While CNNs are commonly thought to be invariant to translation, the gain in performance from \texttt{T.(30\%)} compared to no augmentation shows that \emph{invariance to translation is not by construction built in common architectures such as ResNets}, consistent with \citet{Engstrom19, Kauderer2018, Zhang2019}. 
The gain in performance from \texttt{T.(30\%)} compared to no augmentation is surprising given the (approximate) translation invariance built in to convolutional architectures such as ResNets.
% , and is consistent with \citet{Engstrom19, Kauderer2018, Zhang2019}. 
% \ari{Can we clarify that this was in synthetic/simplistic settings?} We come back to this point when we study models' invariance in Sec. \ref{sec:invariance}. 
Furthermore, \texttt{T.(30\%)} performs almost as well as \texttt{T.(30\%) + RandomSizeCenterCrop}, which has scale augmentation as well. Thus, adding the change in scale to the translation does not further improve performance, which was not the case when comparing \texttt{FixedSizeRandomCrop} and \texttt{RandomSizeCenterCrop}. We compare the validation images that become correctly classified (compared to no augmentation) when using \texttt{FixedSizeRandomCrop}, \texttt{T.(30\%)} and \texttt{T.(30\%) + RandomSizeCenterCrop} in Appendix \ref{sec:appdataaug} but no clear pattern emerged. 

\vspace{-1mm}
\subsection{Trade-off between variance and magnitude of augmentation}
\label{sec:distrib}

What role does the distribution over augmentations in \texttt{RandomResizedCrop} play? Default values for the distribution over $s$ are $s_{-}=0.08,s_{+}=1$. Thus the scale factor can increase the size of an object in the crop up to to $1/s=1/0.08\approx12.5$ times larger. Does only the range of augmentation magnitudes matter? What if we change the shape of this distribution, for the same range of values?

To test this, we modified \texttt{RandomResizedCrop} to use a Beta distribution $B(\alpha,\beta)$ over the standard interval for $s$ ($[0.08,1]$). Fixing $\alpha=1$, we vary $\beta$, which changes the distribution shape. Setting $\beta = 1$ corresponds to a uniform distribution $U(0.08, 1)$, while smaller values of $\beta$ lead to heavily sampling values of $s$ near 1 (and vice versa; see Appendix Fig. \ref{fig:distrib} for visualizations). 
Table \ref{tab:CCTcomp} shows that just varying the shape of the distribution results in a $6\%$ drop in performance. 
To explain this drop, we examine the discrepancy between average apparent object sizes during training and evaluation as in \citet{Touvron2019}. We note smaller values of $\beta$ reduce the discrepancy between image sizes during training and evaluation (see Appendix \ref{sec:appdistrib} for the mean of $B(\alpha,\beta)$). However, very small values of $\beta$ (e.g., 0.1) also decrease performance, as they do not encourage scale invariance by sampling $s$ near 1 (no scale change) and $1/s$ has very little variance. Thus, we observe a trade-off between variability to induce invariance and consistency between training and evaluation object sizes.
% \section{Understanding the extent and source of invariance across architectures and augmentations} 
% \vspace{-3mm}
\section{Invariance across architectures and augmentations}
\label{sec:invariance}
\vspace{-2mm}

So far, we have measured the impact of decomposed augmentations on model performance, but how do these augmentations impact invariance to these and other transformations? To what extent do other elements, such as architectural inductive bias and learning contribute to these invariances? And finally, how do these invariances differ across categories of transformations? In this section, we address these questions by defining a metric for invariance and evaluate this metric for a number of transformations across combinations of architectures, augmentations, and training.

\subsection{Measuring invariance}
A model $f$ is considered invariant to a transformation with a specific magnitude $T_\theta$ if applying $T_\theta$ leaves the output unchanged. We choose to measure invariance by measuring the cosine distance, $d$ between the embeddings of a sample $x$ and its transformed version, relative to a baseline, $b$:
\begin{equation}
    \text{Inv}_{T_{\theta}}(f(x))=\dfrac{b - d(f(x),f(T_\theta(x))}{b}
\end{equation}
where $f$ generates the embedding up from the penultimate layer of a ResNet18 or a vision transformer trained on ImageNet \citep{dosovitskiy2020image,rw2019timm}, see Appendix \ref{sec:appInvariance} for training details. The baseline, $b$, is the embedding distance across two randomly selected samples, $b=d(f(x_i),f(T_\theta(x_j)))$, to account for the effect different transformations (and magnitudes) may have on the distribution of embeddings. The closer $\text{Inv}_{T_{\theta}}$ to $1$, the more invariance to $T_{\theta}$. We report the distribution of $\text{Inv}_{T_{\theta}}$ across pairs.

To measure how invariance changes as transformations intensify, we report invariance as a function of transformation magnitude, with magnitude 0 meaning no transformation and magnitude 9 corresponding to a large transformation (see Appendix Fig. \ref{fig:transforms} for examples). We expect invariance to decrease as transformation magnitude increases for the majority of settings.

\subsection{Invariance to geometric and appearance transformations}
To understand the extent and source of invariance, we measured invariance to common appearance and geometric transformations for both ResNet18 and ViT models. Appearance transformations, such as changes in brightness, alter color or illumination, while geometric transformations such as scaling, translation, and rotation alter the spatial arrangement of pixels.

\begin{figure}[t]
        \centering
    \begin{subfigure}[b]{\textwidth}
        \centering
        \includegraphics[width=\textwidth]{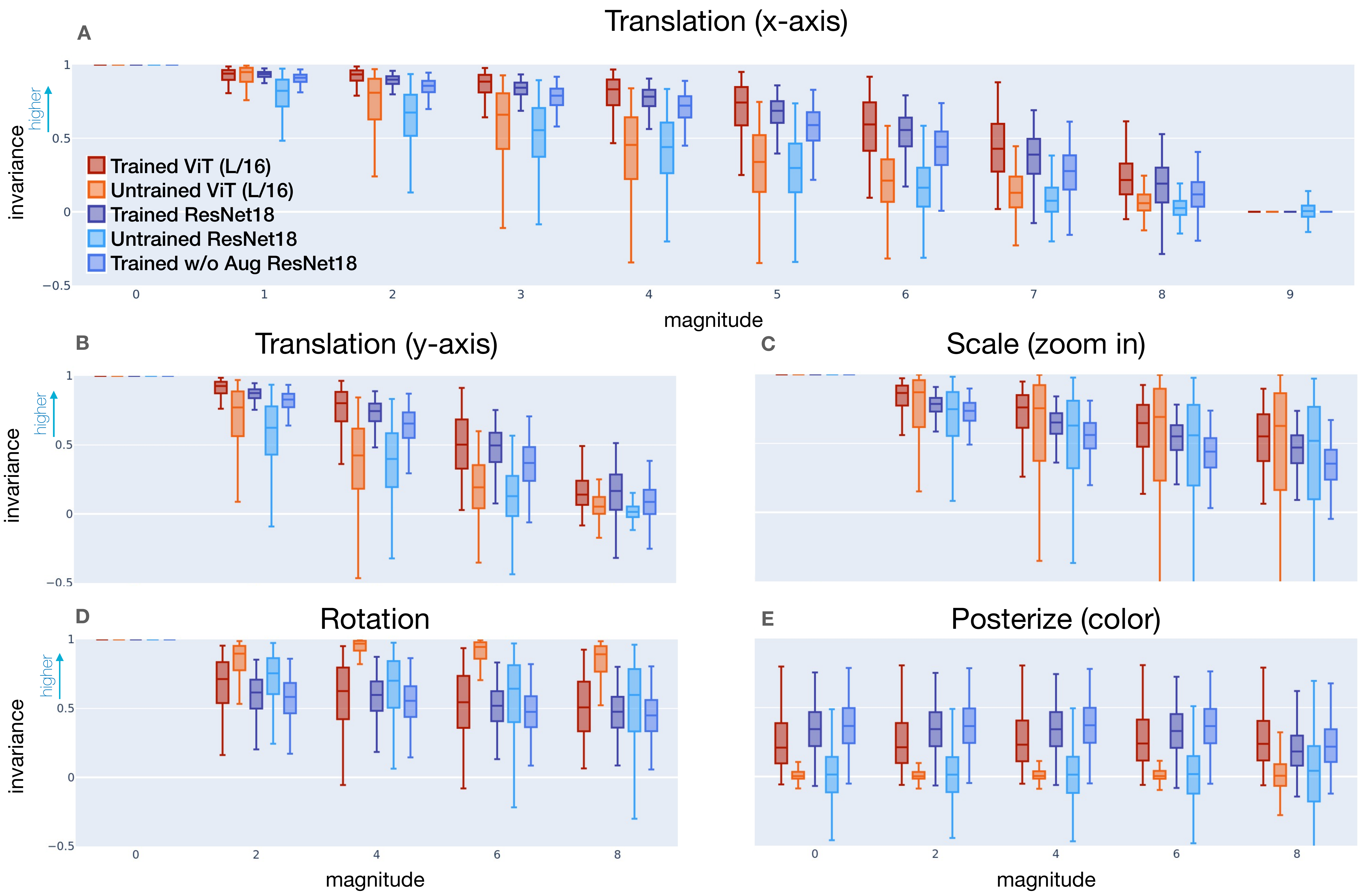}
    \end{subfigure}
    \caption{\textbf{Comparing sources of invariance}. We compare invariance across trained and untrained ResNet18 and ViT to isolate the effect of architectural bias, training, and data augmentation on invariance. Standard augmentations used for ResNet18 and ViT are RRC+horizontal flips. Values are for training pairs, though we found similar trends on validation pairs.}
    \label{fig:geoInvariance}
\end{figure}

First, we found that models indeed featured invariance to a number of common transformations, including translation and scale (Fig. \ref{fig:geoInvariance}). Examining the impact of architecture, we observed that for translation, both ResNet18 and ViT models learned to be invariant, with ViT models consistently achieving slightly higher translation invariance than ResNet18. Surprisingly, untrained ViT models also featured stronger invariance to translation when compared to untrained ResNet18 (Fig. \ref{fig:geoInvariance}a, b, compare orange with light blue), suggesting that despite the convolutional inductive bias present in ResNet18, translation invariance is more prominent in ViT models. 

We next examined the impact of training on invariance. While training consistently increased invariance to translation and to appearance based transformations (Fig. \ref{fig:geoInvariance}a,b and additional figures in Appendix Fig. \ref{fig:appSimChange}), training resulted in no change in average invariance to scale (zooming-in) but reduced the variability in this invariance (Fig. \ref{fig:geoInvariance}). In contrast, training \textit{reduced} invariance to rotation and shear (Fig. \ref{fig:geoInvariance}d and \ref{fig:appSimChange}). Finally, we examined the role of augmentation in learning invariance. Consistent with our finding in Sec. \ref{sec:dataaug}, we found standard augmentation improves translation invariance (albeit only slightly; compare dark and medium blue in Fig. \ref{fig:geoInvariance}a,b). Surprisingly, it barely increases scale invariance, and had equivocal effects on other transformations. Together, these results suggest that the data itself is the major factor influencing learned invariance rather than the choice of architecture or the specific augmentations used. 

We have shown that ResNets present less invariance to translation than vision transformers, despite the inductive bias of the convolutional architecture. As such, what role does the architectural inductive bias in ResNets play? To answer this, we measure equivariance to assess whether models encode predictable responses to transformations. As described in Section \ref{sec:intro}, equivariance is a more general property than invariance: Invariance is a specific case of equivariance and equivariance does not necessarily imply invariance (though invariance does imply equivariance). We evaluate equivariance by examining the alignment of embedding responses to transformations. 
%We find equivariance to translation for untrained ResNet18 that is absent for ViT, highlighting the architectural inductive bias of CNNs to translation 
We find while an untrained ResNet18 is not invariant to translation, it is equivariant to translation, highlighting the architectural inductive bias of CNNs to translation (see Appendix \ref{sec:appInvariance} for detailed results). 

Nevertheless, we observed that trained ViT and ResNet18 are both able to learn invariance to both geometric and appearance transformations, regardless of their particular inductive biases. However, it remains unclear how these learned invariances relate to the factors of variation present in ImageNet. In the next section, we attempt to answer this question.

\section{Characterizing factors of variation with similarity search}
\label{sec:simsearch2}

In the previous section, we assessed the extent and source of invariance to transformations for ResNet18 and ViT.  Among learning, data augmentation, and architectural inductive biases, we identified learning from data as the predominant source of invariance. Consequently, here we investigate which aspects of the data's variation drive models' learned invariances. We characterize the variation in ImageNet in terms of transformations and relate the variation to the models' learned invariances. To do so, we design a metric comparing how well each transformation allows us to travel from one image to another image with the same label, thus capturing a factor of variation.

In the previous section, we assessed whether ResNet18 and ViT are invariant to a set of transformations, and explored to what extent learning, data augmentation and architectural inductive biases impact this invariance. But do learned invariances correspond to the transformations that actually affect factors of variation in the data? What are the transformations that explain variations in ImageNet? In this section, we design a metric to answer these questions by comparing how well each transformation allows us to travel from one image to another image with the same label, thus capturing a factor of variation. Finally, we relate our findings to models' invariances from the previous section.

Characterizing the factors of variation present in natural images is challenging since we don't have access to a generative model of these images. Here, we introduce a metric to determine these transformations based on a simple idea: factors of variation in the data should be able to explain the differences between images of the same class. For example, suppose one of the primary factors of variation in a dataset of animals is pose. In this case, by modifying the pose of one image of a dog, we should be able to match another image of a dog with a different pose. Concretely, we measure the change in similarity a transformation brings to image pairs. For an image pair $(x_1, x_2)$ from the same class and a transformation $T_\theta$ we measure the percent similarity change as 
\begin{equation} \label{eq:simchange}
 \text{SimChange}_{T_\theta} = \frac{sim(f(x_1),f( T_\theta(x_2))) - sim(f(x_1), f(x_2))}{sim(f(x_1), f(x_2))}
\end{equation}
where $sim$ measures cosine similarity (see Appendix \ref{sec:appSimSearch}). To control for any effect from data augmentation, we take $f$ to be a ResNet18 model up to the penultimate layer trained without data augmentation\footnote{Note that if $f$ is fully invariant to $T_{\theta}$ then $\text{SimChange}_{T_\theta}$ will be zero. Thus we also chose $f$ trained w/o augmentation to reduce translation invariance and confirmed full invariance is not achieved in Sec. \ref{sec:invariance}.}
% \footnote{This also reduces the impact the invariances of $f$ have on our similarity metric: e.g. if $f$ is fully invariant to $T_{\theta}$ then $\text{SimChange}_{T_\theta}$ will be zero. Our analysis from Sec. \ref{sec:invariance} comforts us that invariance is not fully achieved.} 
and report values on training image pairs. A higher similarity $\text{SimChange}_{T_\theta}$ among pairs implies a stronger correspondence between the transformation and factors of variation. We select a relevant pool of transformations using AutoAugment, an automated augmentation search procedure across several image datasets \citep{cubuk2019autoaugment}. The set of transformations encompasses geometric and appearance transformations of varying magnitudes\footnote{See Appendix \ref{fig:transforms} for full list of possible transformations.}. For each $T_\theta$ we report $\text{SimChange}_{T_\theta}$ averaged on image pairs. 

This metric has both advantages and disadvantages. Its primary advantage is that it can be measured on realistic image datasets such as ImageNet without requiring access to a ground-truth generative model, as is often used in synthetic datasets. It also uses a pool of possible transformations which are realizable with standard data augmentation techniques, and encompasses both geometric and appearance based transformations. However, because this metric does not exploit information about the generative process, it has limited ability to capture realistic transformations that occur in the abstract space describing semantic image content. Furthermore, it is difficult to make conclusions based on absolute metrics; as such, we use relative comparisons.

\subsection{Can image pairs across ImageNet be described by a consistent set of transformations?} \label{sec:simpairs}
% \MI{Alternative Section Title: Do image pairs vary consistently throughout ImageNet?}

\begin{figure}[t]
        \centering
    \begin{subfigure}[b]{\textwidth}
        \centering
        \includegraphics[width=\textwidth]{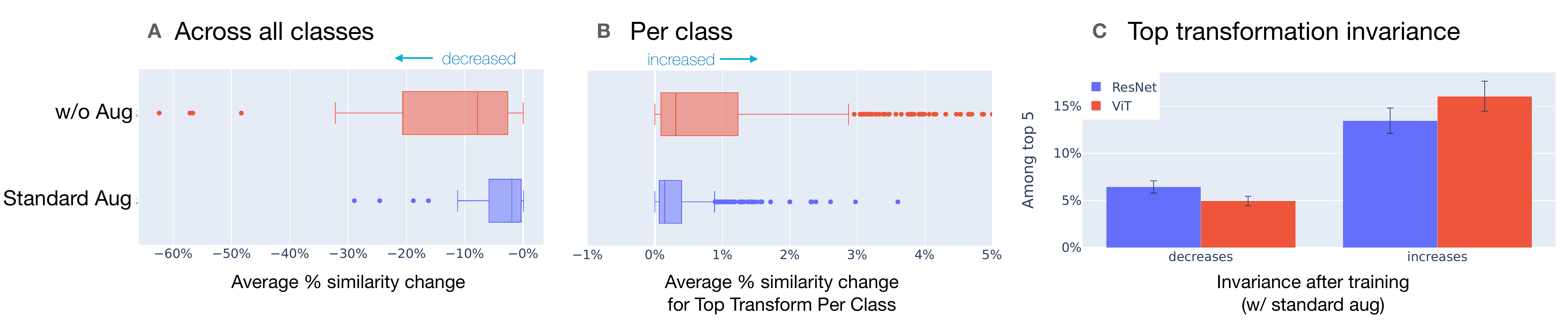}
    \end{subfigure}
    \caption{\textbf{Factors of variation are class-specific and have higher invariance after training}. Fig A. Compares the average percent change in similarity of pairs across all classes. No single geometric or appearance  transformation increases pair similarity across all classes. In contrast, Fig. B shows per class transformations consistently increase similarity. Fig C. shows training increases invariance for transformations more likely to appear among the top 5 transformations per class.}
    \label{fig:simSearchAllClasses}
\end{figure}

Underlying the common practice of data augmentation is the assumption that images ought to vary consistently across a dataset. Throughout training, samples are augmented with the same set of transformations, albeit with differing magnitudes. To test this assumption we ask whether we can explain the variation among image pairs with the same set of transformations. If so, we expect to find a set of transformations which consistently increases the similarity of pairs.  In Fig. \ref{fig:simSearchAllClasses}A we show the distribution of average similarity changes across each transformation as measured by Equation \ref{eq:simchange}. We observe that no single transformation consistently increases average similarity of image pairs across all classes, including geometric and appearance transformations. We find the same pattern holds whether a ResNet18 is trained with or without standard augmentations (RRC + horizontal flips).
% Are there any transformations which consistently increase the similarity between image pairs across all classes? In Fig. \ref{fig:simSearchAllClasses}A we show the distribution of average similarity changes across each transformation. We observe no transformation increases average similarity of image pairs across all classes, including geometric and appearance transformations. We find the same pattern holds whether a ResNet18 is trained with or without standard augmentations (RRC + horizontal flips). 
% When we isolate geometric transformations used in standard augmentations\DB{that are?scale and translation only or all geometric?}, we also find they also do not increase average similarity. 
This result suggests that although standard approaches to data augmentation apply the same transformation distribution to all classes, no single transformation consistently improves similarity (see Appendix \ref{sec:appSimSearch}). Could we instead consistently increase similarity among image pairs by a combination of transformations?

\paragraph{Combinations of transformations do not consistently increase similarity.} 
We repeat our analysis of transformations' effect on image pair similarity using sub-policies,  which combine two transformations of varying magnitudes. We found that while sub-policies can help or hurt by a larger margin as they apply multiple transformations, no sub-policy increases average similarity across all classes (Fig. \ref{fig:appSimSearchSubpolicies}).

\paragraph{Does using combinations of transformations increase similarity?} To answer this, we repeat the analysis using sub-policies, which combine two transformations of varying magnitudes. We found that while sub-policies can help or hurt by a larger margin as they apply multiple transformations, no sub-policy increases average similarity across all classes (Fig. \ref{fig:appSimSearchSubpolicies}).  
% Although, we did not test combinations of transformations beyond sub-policies, we hypothesize the lack of universality is instead due to intrinsic difference among factors of variation by class, which we study in Sec. \ref{sec:SimSearchPerClass}. For example, we expect rotation to be a factor of variation for a dog rolling around, but not a building that's likely to appear straight.

Counter to the common practice of data augmentation, this result suggests the same set of transformations does not consistently explain the variation among image pairs. 
What effect then, if any, does augmenting with the same set of standard transformations have on image similarity?

\paragraph{Training augmentations dampen pair similarity changes.} Similar to our earlier results, we observed that training with augmentations induce more invariance relative to training without augmentations. Though all transformations consistently decrease similarity for models trained with and without augmentation, models trained with augmentation exhibited both a smaller decrease in similarity and lower variance (Fig. \ref{fig:simSearchAllClasses}A). Training with augmentation also reduced the decrease in similarity for transformations beyond simply scale and translation (Appendix Fig. \ref{fig:simSearchNotStandard}), suggesting that these augmentations impact the response even to transformations not used during training.

% This results suggests that standard augmentation decrease the magnitude by which transformation affect embedding changes. We also verify whether standard augmentation do so for other transformations in Appendix \ref{sec:appSimSearch} by measuring the distribution of average similarity changes after excluding translation and scale. This suggests standard augmentations affect the responses of embeddings even to transformations not used during training augmentation.
% \MI{Consider breaking up analysis by transformation to introduce nuance around which transformations are most effected}\DB{Link with comparison of invariance properties of standard versus no aug model}

\subsection{Are factors of variation specific to each class?}
\label{sec:SimSearchPerClass}

In Sec. \ref{sec:simpairs}, we showed no transformation (or sub-policy) consistently increased similarity across classes. However, the factors of variation and consequently, the optimal transformation, may be different for different classes, especially those which are highly different. Can we instead consistently increase similarity if we allow flexibility for transformations to be class specific?

To test this hypothesis, we examined the top transformation for each class. In contrast to the global result, we found the top transformation per class consistently increased the average similarity for all classes (Fig. \ref{fig:simSearchAllClasses}B). Notably, the top transformation per class increased similarity by $3.36 \pm 0.9\%$ (mean $\pm$ SEM) compared to effectively no change ($0.020\% \pm 0.021\%$) for the top transformation across all classes. Similar to \citet{Hauberg16} which learn per class transformations for data augmentation that boost classification performance on MNIST \citep{lecun2010mnist}, we applied per class data augmentation variants on ImageNet, but observed no significant classification performance boost. We leave further applications of per class augmentation to future work. 

\paragraph{Data augmentation is not beneficial for all classes.} Since the optimal transformation differs across classes, do standard augmentations benefit all classes or only some classes? To test this, we examined the impact of RRC on the performance of individual ImageNet classes. Interestingly, we found that on average\footnote{Computed over 25 pairwise comparisons of 5 runs with both augmentations.} $12.3 \pm 0.21\%$ have a lower performance when using RRC. Critically, some classes are consistently hurt by the use of RRC with a difference in top-1 accuracy as high as $22\%$. We systematically study these classes in Appendix \ref{sec:apphurt}, but no clear pattern emerged.

\subsection{Appearance transformations are more prevalent} 

Data augmentation and most of the literature on invariant models often rely only on geometric transformations such as translations, rotations, and scaling.
% As we explained, \texttt{RandomResizedCrop} combines both translation and scale for example.
However, it is not clear whether or not the factors of variation in natural images are primarily geometric. If this is the case, we would expect the top transformations from our similarity search to be geometric rather than appearance-based. In contrast, we found that appearance transformations accounted for more variation in ImageNet than geometric transformations, consistent with recent work \citep{cubuk2019autoaugment}. Of the top transforms, $78\%$ were appearance based compared to only $22\%$ geometric. We confirmed this difference is not due to a sampling bias by ensuring an approximately even split between geometric and appearance transformations. In fact, if we isolate geometric transformations, we find for $64\%$ of classes the top transformation is the identity, suggesting geometric transformations are worse when applied to an entire class than no transformation at all. We find a similar pattern among the top transforms per class for ResNet18 trained with standard augmentations: for more than $98.4\%$ of classes the top transformation alters appearance not geometry. In Appendix \ref{sec:appLocalVariation} we also examine local variation in foregrounds to translation and scale. 

% We corroborate results suggested by earlier work: appearance transformations account for more variation in ImageNet compared to geometric \citep{cubuk2019autoaugment}. Table \ref{tab:simSearchGeo} shows the number of times a given transformation appears as the top transformation for each class, measured by the average percent increase in similarity. We also show an alternative ranking of top transformations by proportion of pairs with increased similarity.  \MI{TODO this sentence is a detail for experimental setup/appendix}. We see that appearance transformations accounting for more than 3x the number of top transformations compared geometric transformations. We confirmed this difference is not due to a sampling bias by ensuring an approximately even split between geometric and appearance transformations. Among the top appearance transformations is posterize along with other appearance-based such as contrast and color. 

\subsection{Training increases invariance for factors of variation}

In Sec. \ref{sec:invariance}, we showed that training increases invariance to a number of transformations, independent of architectural bias and augmentation and in Sec. \ref{sec:simpairs}, we used similarity search to characterize the transformations present in natural images. However, do the invariances learned over training correspond to the factors of variation present in natural images? 

To test this, we asked whether the transformations to which networks learn to be invariance correspond to the same transformations which are highly ranked in similarity search. We found transformations that exhibit increased invariance over training were substantially more likely to be ranked in the top 5 transformations per class compared to transformations which exhibited decreased or minimal change in invariance over training (Fig. \ref{fig:simSearchAllClasses}C). This result demonstrates training increases invariance to factors of variation present in real data, regardless of whether there is an inductive bias or data augmentation is specifically designed to encourage invariance. 

\subsection{Characterizing factors of variation across classes}

We have shown that there exist factors of variation which consistently increase similarity for a given class, but it remains unclear why a particular factor might impact a particular class. Here, we investigate whether related classes feature related factors of variation. 
% Although, a transformation which increases the similarity of a single pair of images is by definition changing a factor of variation for the pair, here we characterize more prominent connections between similarity changes and factors of variation.
% We first consider the most clear cut cases by ranking top transformations by weighted boost, defined as proportion of images with an increase in similarity multiplied by the average percent increase. We expect transformations with a high weighted boost to corresp  d a factor of variation for its corresponding class. 

One prominent pattern which emerged was among textile-like classes such as velvet, wool, handkerchief, and envelope. When considering single transformations, rotation is the top transformation or rotation plus an appearance transformations (such as color or posterize) for sub-policies (see Appendix \ref{app:simSearchWeightedBoost} for a full list). The relationship between rotation and textiles makes intuitive sense: fabrics generally don't have a canonical orientation and can appear in many different colors. 

\begin{figure}[t]
    \centering
    \includegraphics[width=\textwidth]{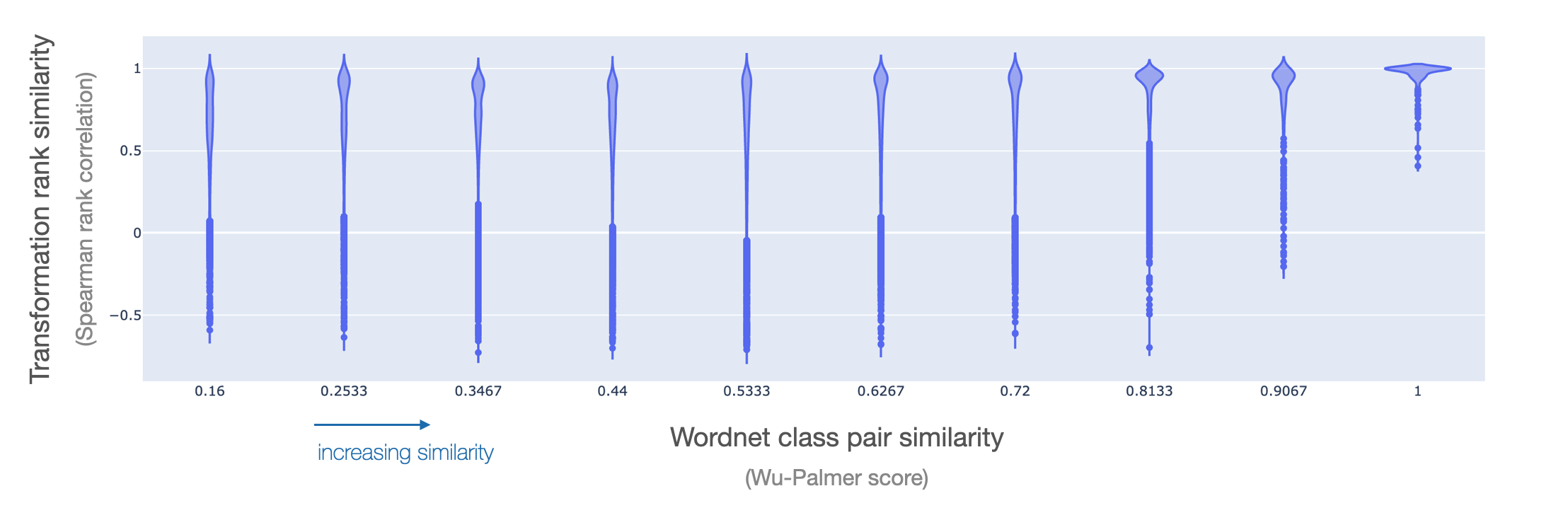}
    \vspace{-2em}
    \caption{\textbf{Similar classes share factors of variation}. Wordnet class pair similarity versus transformation ranking. Violin plots show the distribution of transformation Spearman's rank correlation as class pair similarity increases. We see similar classes rank transformations similarly.}
    \label{fig:wordnetSimSearch}
\end{figure}

To test this systematically, we measured the pairwise class similarity using Wordnet \citep{wordnet} and compared it to the similarity between the top transforms for each class. We computed class similarities using the most specific common ancestor in the Wordnet tree against the Spearman rank of transformation types. We found that while dissimilar classes often have similar transformations, similar classes consistently exhibit more similar transformations (Fig. \ref{fig:wordnetSimSearch}; Appendix \ref{sec:appSimSearchWordnet}).

\section{Related work}
\paragraph{Data augmentation approaches.} Standard data augmentations often amount to taking a crop of random size and location, horizontal flipping and color jittering \citep{He2015, Touvron2019, Simonyan15}. In self-supervised learning, \citep{NEURIPS2020_70feb62b} boost performance by using multiple crops of the same image. \citet{Hauberg16} learn per class augmentation and improve performance on the MNIST dataset \citep{lecun2010mnist}. AutoAugment is a reinforcement learning-based technique that discovers data augmentations that most aid downstream performance \cite{cubuk2019autoaugment}. \citet{antoniou2018data} train a Generative Adversarial Networks (GANs \citep{NIPS2014_5ca3e9b1}) based model to generate new training samples. \citet{Wilk2018} follow a Bayesian approach and integrate invariance into the prior.
% , the model is then learned using marginal likelihood maximization. 
Recent works aim to automatically discover symmetries in a dataset \citep{Hashimoto2017}, and enforce equivariance or invariance to these \citep{benton2020learning,zhou_meta-learning_2020,dehmamy2021lie}. While these methods are promising, they have mostly been applied to synthetic datasets or augmented versions of real datasets. Their application to a real dataset such as ImageNet is not straightforward: we tried applying the Augerino model \citep{benton2020learning} to ImageNet, we found it was struggling to discover effective augmentations composed of multiple transformations (see Appendix \ref{sec:augerino}).

\paragraph{Consistency of neural architectures.}\citet{Zhang2019} show that invariance to translation is lost in neural networks, and propose a solution based on anti-aliasing by low-pass filtering. \citet{Kauderer2018} studies the source of CNNs translation invariance of CNNs on a translated MNIST dataset \citep{lecun2010mnist} using translation-sensitivity maps based on Euclidean distance of embeddings, and find that data augmentation has a bigger effect on translation invariance than architectural choices. Very recently, \citet{Myburgh2021} replace Euclidean distance with cosine similarity, and find that fully connected layers contribute more to translation invariance than convolutional ones.
% In a similar vein, \citet{lenc18understanding} show AlexNet has invariance to translation in early layers by examining horizontal flip, rescaling by half, and rotation by 90 degrees. 
\citet{lenc18understanding} study the equivariance, invariance and equivalence of different convolutional architectures with respect to geometric transformations. Here, we study invariance on a larger set of transformations with varying magnitudes and compare ResNet18 and ViT. \citet{Touvron2019, Engstrom19} explore the specificities of standard data augmentations, and \citet{Engstrom19} found that a model augmented at training with rotations and translations still fails on augmented test images.
% Indeed, while data augmentation is expected to bring invariance to a set of transformations,
% such as scaling and translation in the case of random resized crop, 
% \citet{Engstrom19} found that a model augmented with rotations and translation still fails when presented with test images augmented by worst-case rotations and translation. 
% Interestingly, they improve robustness by using a robust optimization scheme and majority voting at inference time. 
We differ from these works by grounding invariance into the natural factors of variation of the data, which we try to characterize. To the best of our knowledge, the links between invariance and the data factors of variation has yet not been studied on a large-scale real images dataset.

\section{Discussion}

In this work, we explored the source of invariance in both convolutional and vision transformer architectures trained for classification on ImageNet, and how these invariances relate to the factors of variation of ImageNet. We compared the impact of data augmentation, architectural bias and the data itself on the models' invariances. We observed that \texttt{RandomResizedCrop} relies on a precise combination of translation and scale that is difficult to reproduce and that, surprisingly, augmenting with translation recaptures most of the improvement despite the (approximate) invariance to translation built in to convolutional architectures. By analyzing the source of invariance in ResNet18 and ViT, we demonstrated that invariance generally stems from the data itself rather than from architectural bias, and is only slightly increased by data augmentation. 

Finally, we connected the models' learned invariance to the factors of variation present in the data. We characterized variation in ImageNet by examining pair similarity in response to transformations, finding that transformations which explain variation in ImageNet are class-specific, more appearance-based, and align with the invariances' learned during training.

% Finally, we showed that the transformations that explain the variations in ImageNet are per class and mostly appearance based. Interestingly, we found invariance and factors of variation align: training enhances a model's natural invariance for transformations that account for ImageNet variations, while it has the opposite effect for changes that do not affect factors of variation. 

\paragraph{Limitations and future work} We provide an analysis of the invariant properties of models using a specific set of metrics based on model performance and similarities of input embeddings. Using these, some of our conclusions are shared with existing works, but a different set of metrics could potentially bring more insights on our conclusions. Additionally, we only experiment on ImageNet, but it would be interesting to perform the same type of analysis on a wider range of datasets and data types. Does a handful of transformations describe the variations of most standard image datasets? Our study sheds light on ImageNet factors of variation but some conclusions remain obscure, such as the role of scaling. This emphasizes the difficulty of performing a systematic study of real datasets. 

Finally, our findings spark further exploration. Could tailoring augmentations per class or introducing appearance-based augmentations improve performance? 

\paragraph{Potential negative societal impacts.} While our work is concerned with robustness of vision models which can have various societal impacts, our work is primarily analytical and we do not propose a new model. Hence, we do not foresee any potential negative societal impacts of our findings. Our study does emphasize the importance of dataset construction, as the predominant source of invariance, relative to other modelling considerations. Consequently, our work encourages researchers to also consider dataset construction as an important aspect of vision models' societal impact.

\paragraph{Acknowledgements} We would like to thank David Lopez-Paz, Armand Joulin, Yann Olivier, Kamalika Chaudhuri, Aaron Aadock, Pascal Vincent, Michael Alcorn, and Nicolas Ballas for helpful discussions and feedback.

\bibliography{diane.bib}
\bibliographystyle{plainnat}

%%%%%%%%%%%%%%%%%%%%%%%%%%%%%%%%%%%%%%%%%%%%%%%%%%%%%%%%%%%%
\newpage
\section*{Checklist}

\begin{enumerate}

\item For all authors...
\begin{enumerate}
  \item Do the main claims made in the abstract and introduction accurately reflect the paper's contributions and scope?
    \answerYes{}
  \item Did you describe the limitations of your work?
    \answerYes{See the Discussion section.}
  \item Did you discuss any potential negative societal impacts of your work?
    \answerYes{See the Discussion section.}
  \item Have you read the ethics review guidelines and ensured that your paper conforms to them?
    \answerYes{}
\end{enumerate}

\item If you are including theoretical results...
\begin{enumerate}
  \item Did you state the full set of assumptions of all theoretical results?
    \answerNA{We did not include theoretical results.}
	\item Did you include complete proofs of all theoretical results?
    \answerNA{We did not include theoretical results.}
\end{enumerate}

\item If you ran experiments...

\begin{enumerate}
  \item Did you include the code, data, and instructions needed to reproduce the main experimental results (either in the supplemental material or as a URL)?
    \answerYes{Code to reproduce the main experiments is available at \url{https://github.com/facebookresearch/grounding-inductive-biases}.}
  \item Did you specify all the training details (e.g., data splits, hyperparameters, how they were chosen)?
    \answerYes{See details in each experiment section, as well as Appendix \ref{sec:trainingdetails}, \ref{sec:augerinodetails}, \ref{sec:augtxtyscale} \ref{sec:appSimSearch}, \ref{sec:appInvariance}.}
	\item Did you report error bars (e.g., with respect to the random seed after running experiments multiple times)?
    \answerYes{We report standard error of the mean (SEM) for all experiments.}
	\item Did you include the total amount of compute and the type of resources used (e.g., type of GPUs, internal cluster, or cloud provider)?
    \answerYes{See Appendix \ref{sec:trainingdetails}.}
\end{enumerate}

\item If you are using existing assets (e.g., code, data, models) or curating/releasing new assets...
\begin{enumerate}
  \item If your work uses existing assets, did you cite the creators?
    \answerYes{We cite all assets and existing code / pre-trained models we used (ImageNet dataset, Wordnet database, U2Net, pre-trained ViT, Augerino library, AutoAugment).}
  \item Did you mention the license of the assets?
    \answerNA{We used existing assets which licenses can be found in the references we cite. While ImageNet does not own the copyright of the images, they allow non-commercial uses of the data under the terms of access written in \texttt{https://image-net.org/download.php}.}
  \item Did you include any new assets either in the supplemental material or as a URL? \answerYes{While we don't provide any new assets such as datasets, we provide code to reproduce our main results in the Supplementary.}
  \item Did you discuss whether and how consent was obtained from people whose data you're using/curating?
    \answerNA{We use the ImageNet dataset as is and reference to it.}
  \item Did you discuss whether the data you are using/curating contains personally identifiable information or offensive content?
     \answerNA{We use the ImageNet dataset as is and reference to it.}
\end{enumerate}

\item If you used crowdsourcing or conducted research with human subjects...
\begin{enumerate}
  \item Did you include the full text of instructions given to participants and screenshots, if applicable?
    \answerNA{We did not conduct research with human subjects or used crowdsourcing.}
  \item Did you describe any potential participant risks, with links to Institutional Review Board (IRB) approvals, if applicable?
    \answerNA{We did not conduct research with human subjects or used crowdsourcing.}
  \item Did you include the estimated hourly wage paid to participants and the total amount spent on participant compensation?
    \answerNA{We did not conduct research with human subjects or used crowdsourcing.}
\end{enumerate}

\end{enumerate}
%%%%%%%%%%%%%%%%%%%%%%%%%%%%%%%%%%%%%%%%%%%%%%%%%%%%%%%%%%%

\newpage
\begin{appendices}
\renewcommand{\thefigure}{A\arabic{figure}}
\setcounter{figure}{0}
\renewcommand\thetable{A\arabic{table}}
\setcounter{table}{0}
\section{Training details}
\label{sec:trainingdetails}

\paragraph{Training regular Resnet-18} 
The experiments of Sec. \ref{sec:dataaug} are conducted using 1 seed to cross-validate between 3 learning rates ($0.01,0.1,0.5)$ and 3 weight decay parameter $0.01,0.001,0.0001$. Models are trained with Stochastic Gradient Descent with momentum equal to 0.9 \citep{journals/nn/Qian99} on all parameters. We use a learning rate annealing scheme, decreasing the learning rate by a factor of 0.1 every 30 epochs. We train all models for 150 epochs. 
Then, we select the best learning rate and weight decay for each method and run 5 different seeds to report mean and standard deviation. We use the validation set of ImageNet to perform cross-validation and report performance on it. Our code is a a modification of the pytorch example found in \texttt{https://github.com/pytorch/examples/tree/master/imagenet}.\\
\\
Note that we also tried one seed with cross-validation of hyper-parameters of \texttt{T.(50\%) + RandomSizeCenterCrop}, i.e. with $50\%$ translation, this gives poorer performances than $30\%$ translation (top-1 accuracy $~\approx 67.5\%$).

Code to reproduce experiments is available at \url{https://github.com/facebookresearch/grounding-inductive-biases}.

\paragraph{Training the Augerino model} 
In section \ref{sec:augerino} we train the Augerino method on top of the Resnet-18 architecture. We employ Augerino on top of applying the \texttt{FixedSizeCenterCrop} pre-processing, in order to not induce any invariance by data augmentations. The experiments reported in section \ref{sec:augerino} are conducted using 5 seeds to cross-validate between 7 regularization values $\lambda$ ($0.01,0.1,0.2,0.4,0.6,0.8,1$). We use the best learning rate and weight decay values of the Resnet-18 trained with \texttt{FixedSizeCenterCrop} (learning rate $0.1$ and weight decay $0.0001$). The parameters of the distribution bounds, specific to Augerino, are trained with a learning rate of $0.01$ and no weight decay as in the original Augerino code (\texttt{https://github.com/g-benton/learning-invariances}). Models are trained with Stochastic Gradient Descent with momentum equal to 0.9 \citep{journals/nn/Qian99} on all parameters. We use a learning rate annealing scheme, decreasing both learning rates of the Resnet-18 and the Augerino bounds parameters by a factor of 0.1 every 30 epochs. We train all models for 150 epochs. During training, 1 copy of the image transformed with the sampled transformation values is used, and during validation and test, 4 transformed versions of the image are used, as in the original Augerino code. We use the validation set of ImageNet to perform cross-validation and report performance on it. 

\paragraph{Total amount of compute, type of resources used} Our main code runs with the following configuration: pytorch 1.8.1+cu111, torchvision 0.9.1+cu111, python 3.9.4. Full list of packages used is released with our code. We use \texttt{DistributedDataParallel} and ran the experiments on 4 GPUs of 480GB memory (NVIDIA GPUs of types P100 and V100) and 20 CPUs, on an internal cluster. With this setting, training a ResNet-18 on ImageNet with \texttt{RandomResizedCrop} takes approximately 6 mins, while for other augmentations (e.g.\texttt{T.(30\%) + RandomSizeCenterCrop}) it can take up to approximately 20 mins depending on the type of GPU used.\\
\\
For the experiments of Sec. \ref{sec:augerino} we use pytorch 1.7.1+cu110, torchvision 0.8.2+cu110, python 3.9.2 and torchdiffeq 0.2.1 as the Augerino original code relies on torchdiffeq and we could not run torchdiffeq with pytorch 1.8.1 (known issue, see \texttt{https://github.com/rtqichen/torchdiffeq/issues/152}).  

\section{Comparing the samples helped by translation and/or scaling}
\label{sec:appdataaug} 

In Section \ref{sec:dataaug}, we find that \texttt{T.(30\%)} performs on par with \texttt{T.(30\%) + RandomSizeCenterCrop}, and both outperform \texttt{FixedSizeRandomCrop}. To further study this, we compute the lists of samples that are incorrectly classified by no augmentation but correctly classified by these methods, for each of the three methods. As we trained $5$ seeds per methods, we have 25 lists of ``helped samples" for each method. We then compare methods using the intersection-over-union (IoU) of their respective lists. For each method, the IoUs of each method's lists with itself are:
\begin{itemize}
    \item \texttt{T.(30\%)}/\texttt{T.(30\%)}: $0.269 \pm 0.007$
    \item\texttt{T.(30\%) + RandomSizeCenterCrop}/\texttt{T.(30\%) + RandomSizeCenterCrop}: $0.293 \pm 0.007$ 
    \item \texttt{FixedSizeRandomCrop}/\texttt{FixedSizeRandomCrop}: $0.242 \pm 0.007$
\end{itemize}
while the cross-methods lists IoU are:
\begin{itemize}
    \item \texttt{FixedSizeRandomCrop}/\texttt{T.(30\%) + RandomSizeCenterCrop}: $0.220 \pm 0.004$
    \item \texttt{FixedSizeRandomCrop}/\texttt{T.(30\%)}: $0.224 \pm 0.005$
    \item \texttt{T.(30\%)+RandomSizeCenterCrop}/\texttt{T.(30\%)}: $0.242 \pm 0.005$
\end{itemize}
Thus, we do not see a pattern of consistency neither in the intra-methods or cross-methods IoUs. 

\section{Varying the distribution over the scale}
\label{sec:appdistrib}
For the experiment in Sec. \ref{sec:distrib}, we used the best learning rate and weight decay ($0.1$ and $0.0001$) found by cross-validation for \texttt{RandomResizedCrop}. We run $5$ training seeds of each distribution.
\begin{figure}[h!]
        \centering
        \includegraphics[width=0.7\textwidth]{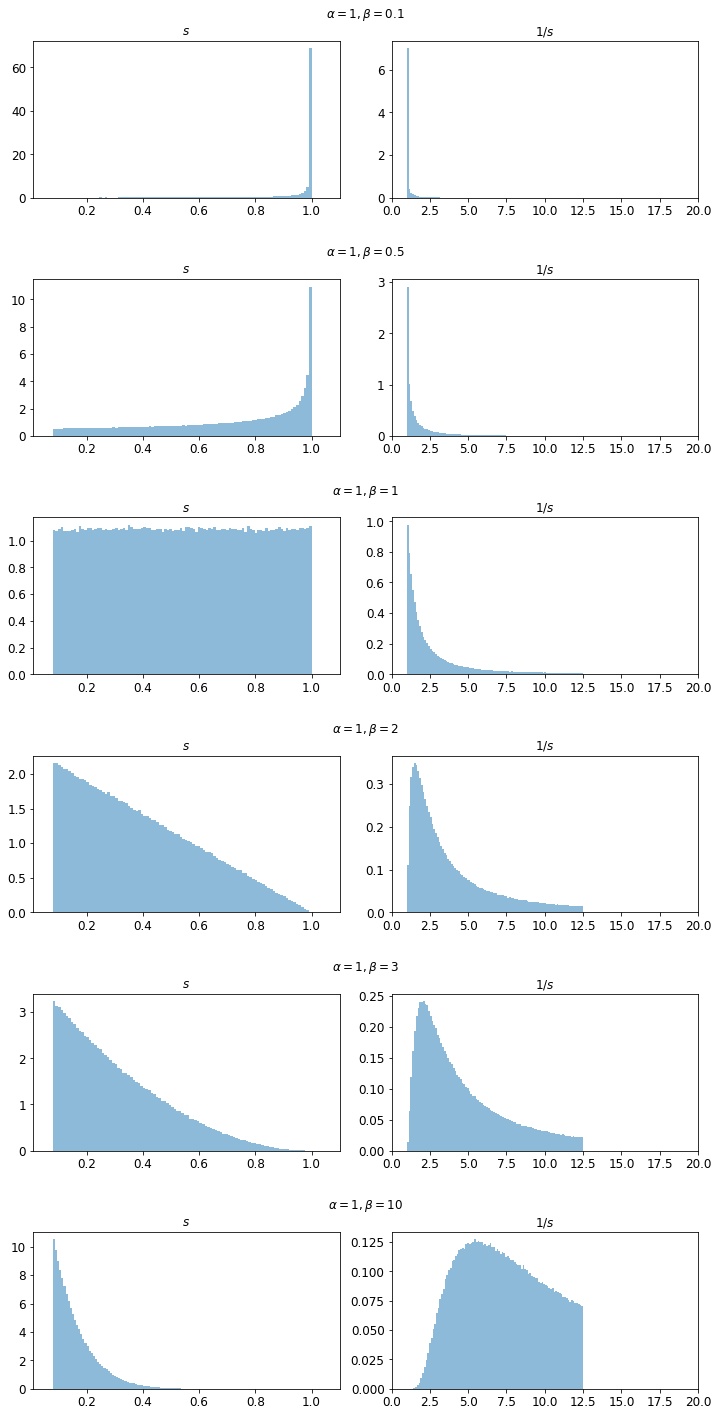}
        \caption{Effect of the value $\beta$ over the distribution of $s$ and $1/s$.}
        \label{fig:distrib}
\end{figure}

The expected value of the non-standard Beta distribution for $\alpha=1$ is $\dfrac{1}{1+\beta}(1-0.08)+0.08$. As explained in \citet{Touvron2019}, using a random scale induces a discrepancy between the average objects apparent sizes during training and evaluation: their ratio is proportional to the expected value of $s$. Thus, the smaller $\beta$ the smaller the expected value of $s$, which reduces this discrepancy.\\
\\
However, too small values of $\beta$ (e.g. $\beta=0.1$) also have smaller performance. Recall that \texttt{RandomResizedCrop} scales an object in the selected crop by a factor proportional to $1/s$. While we don't have a closed form expression for $\mathbb{V}[1/s]$, the variance of $1/s$ (inverse of the non-standard Beta over $s$), we computed estimated values for $\mathbb{V}[1/s]$ using $500 000$ samples drawn from $s$. \footnote{We also checked the estimated values were sensible obtained by computing the probability distribution function (pdf) of $1/s$ from the pdf of $s$ via the change of variable formula, and compute $\mathbb{V}[1/s] = \mathbb{E}[1/s^2]-\mathbb{E}[1/s]^2$ using Wolfram Alpha \citep{wolfram}.}
\begin{table}[h!]
\begin{tabular}[t]{c|c}
      $\beta$ & $\mathbb{V}[1/s]$\\
      \hline
        0.1 & 0.823\\
        0.5 & 3.193\\
        1 ($\sim$ RRC) & 4.962\\
        2 & 6.809\\
        3 & 7.626\\
        10 & 7.317\\
    \end{tabular}
    \caption{Variance of $1/s$.}
    \label{tab:varbeta}
\end{table}
Table \ref{tab:varbeta} shows that the variance for $\beta=0.1$ is $\approx$ 3 and 5 times smaller than the variance for $\beta=0.5$ and $\beta=1$ respectively. Thus, while the three values of $\beta$ give distributions that peak close to $1/s=1$, the value $\beta=0.1$ gives a smaller variance thus less encourage scale invariance. This in our view explains the poorer performance of $\beta=0.1$.
% The variance of the non-standard beta distribution for $\alpha=1$ is $(1-0.08)^2\dfrac{\beta}{(1+\beta)^2(\beta+2)}$.

\section{Invariance and Equivariance}
\label{sec:appInvariance}

\paragraph{Experimental details} We use a pretrained ViT (L/16) and untrained ViT from \cite{rw2019timm} using the `forward features' method to generate embeddings, see Timm's documentation for details \url{https://rwightman.github.io/pytorch-image-models/feature_extraction/}. The trained ViT achieves $80\%$ Top 1 accuracy on ImagNet. For ResNet18, we use the pretrained ResNet18 available from PyTorch \url{https://pytorch.org/hub/pytorch_vision_resnet/}. The trained ResNet18 achieves $70\%$ Top 1 accuracy. For all our measures, we control for the difference in test-set versus train-set sizes by limiting the total number of embedding comparisons to $10k$ pairs.

\begin{figure}[h!]
        \centering
        \includegraphics[width=\textwidth]{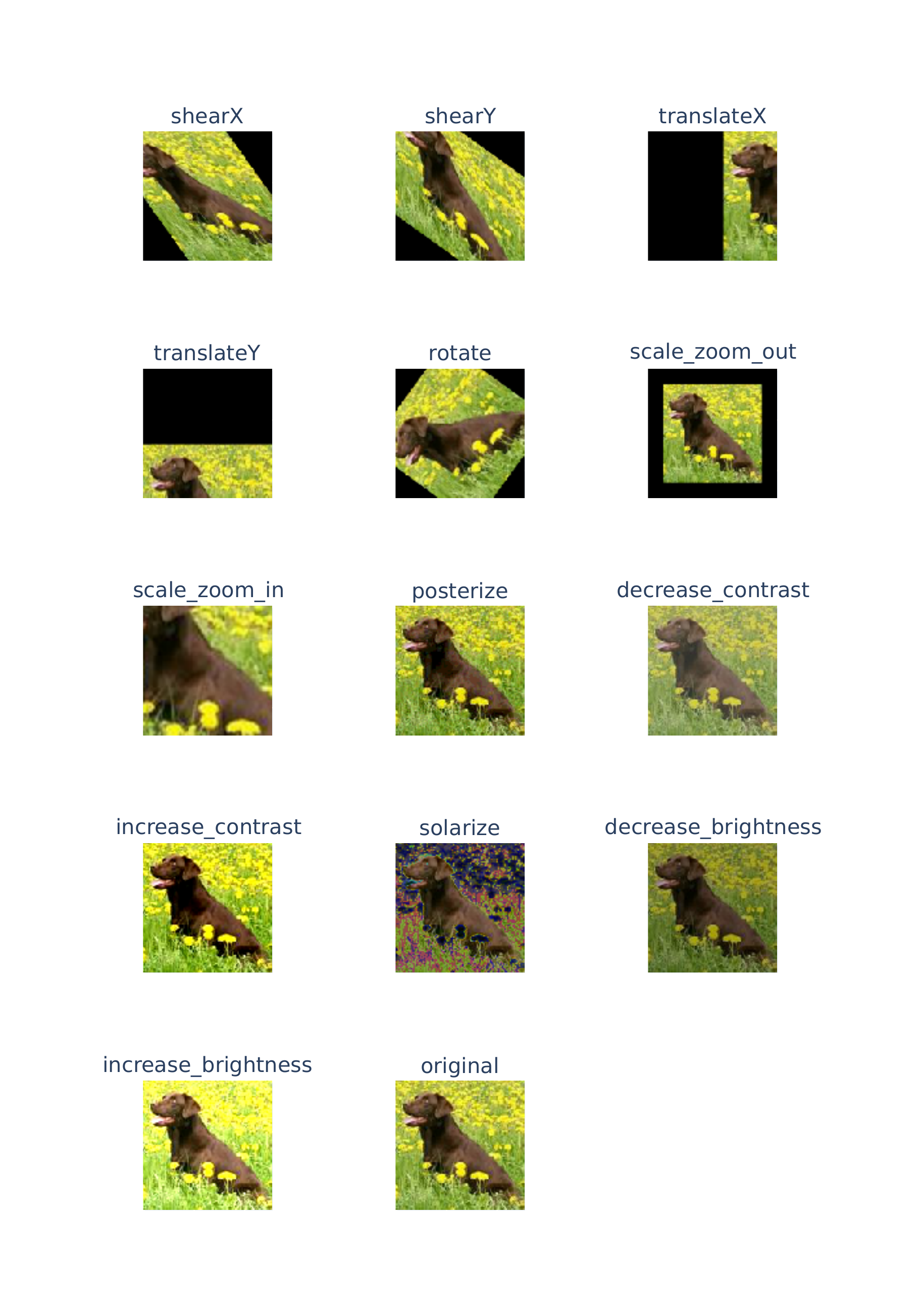}
        \caption{Illustrates the effect of each transformation. The image is from Wikimedia Commons \url{https://commons.wikimedia.org/wiki/File:Labrador_Chocolate.jpg} under Creative Commons Attribution-Share Alike 3.0 Unported license.}
        \label{fig:transforms}
\end{figure}

\begin{figure}[h!]
        \centering
        \includegraphics[width=\textwidth]{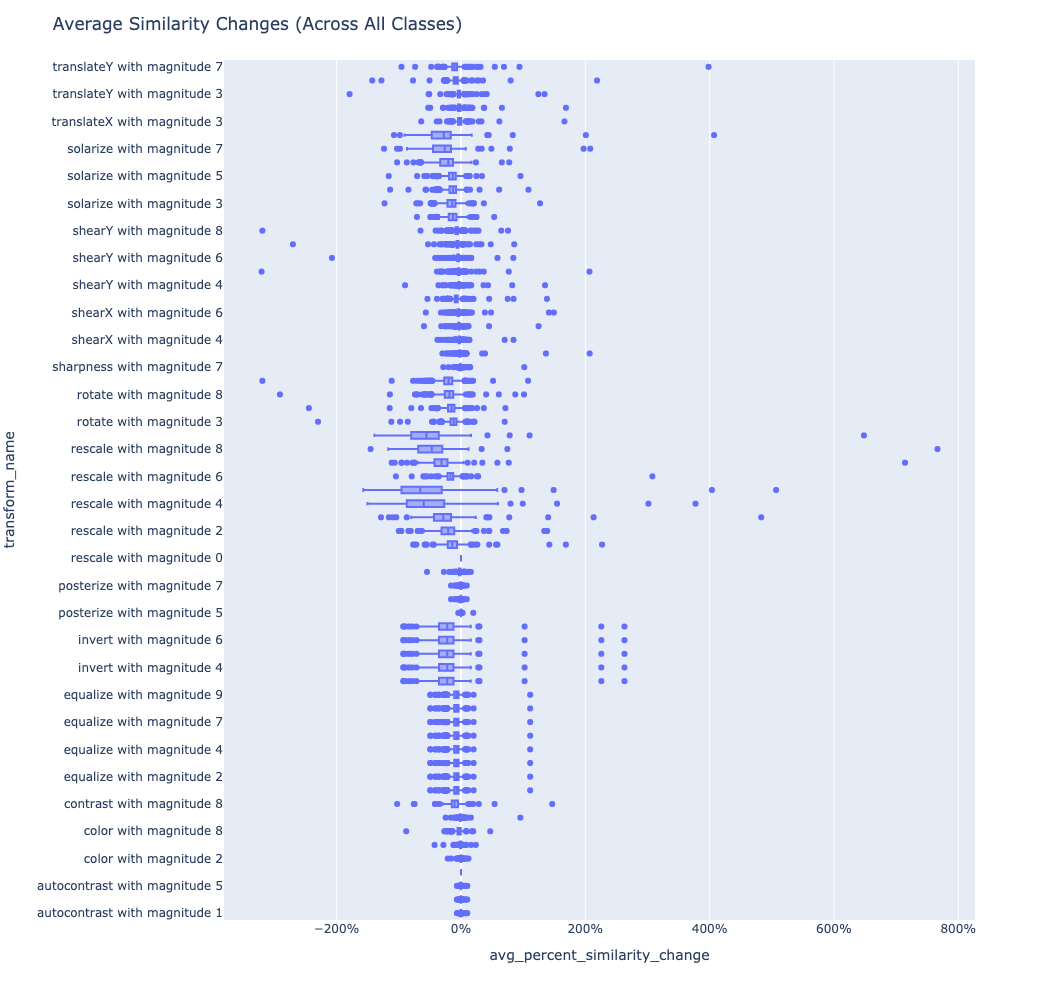} \caption{Similarity Changes by transformation across all classes measured on ImageNet training samples (we find a similar pattern on the validation set). We see no transformation similarity change distribution are around or below zero. } 
        \label{fig:appSimChange}
\end{figure}

\paragraph{Equivariance}
We also explore whether models are equivariant to transformations. A model is said to be equivariant if it responds predictably to the given transformation. Formally, a model, $f$, is equivariant to a transformation $T_\theta$ of an input $x$ if the model's output transforms in a corresponding manner via an output transformation $T_\theta'$, i.e. $T'_\theta(f(x)) = f(T_\theta(x))$ for any $x$. 

\begin{figure}[h!]
    \centering
    \includegraphics[width=\textwidth]{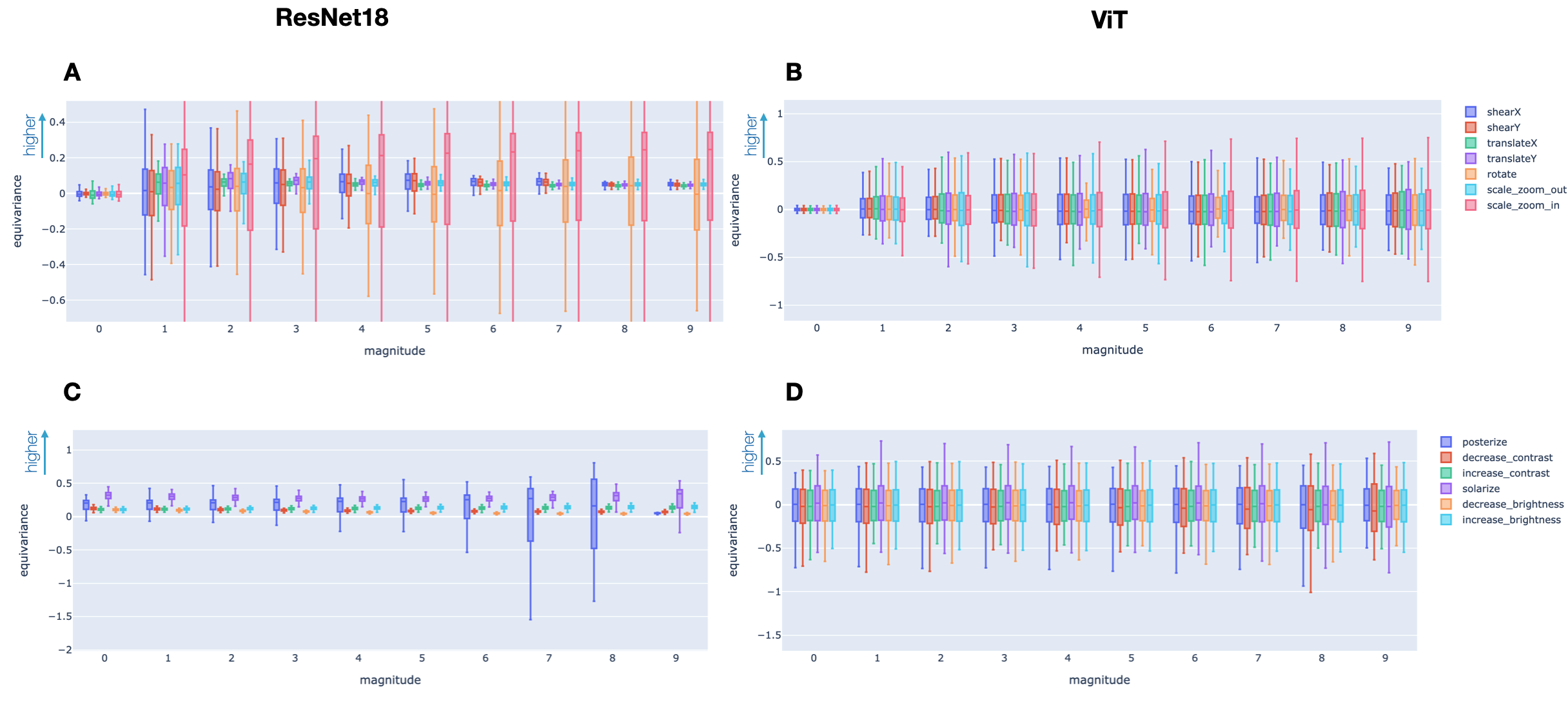}
    \caption{Equivariance for untrained ResNet18 (panel A, C) and ViT (panel B, D). We compare both geometric and appearance transformations using validation set images. We find no significant difference between training and validation set results.}
    \label{fig:equivariance}
\end{figure}

To disambiguate invariance from equivariance, we measure equivariance by examining whether embeddings respond predictably to a given transformation. To do so, we measure alignment among embedding differences, by first producing embedding differences

$$
d_i = f(x_i) - f(T_\theta(x_i))
$$

then measuring pairwise alignment of the embedding differences via cosine similarity. We compare, $sim(d_i, d_j)$ against a baseline B where we shuffle the rows in each column independently. We report $sim(b_i, b_j) - sim(d_i, d_j)$, where $b$ are elements from the baseline. A higher value implies higher equivariance, with $0$ indicating no equivariance above the baseline. In Fig. \ref{fig:equivariance}, we find equivariance to translation for untrained ResNet18 that is absent for ViT, highlighting the architectural inductive bias of CNNs to translation. Although for some magnitudes we also observe equivariance to zooming out, we note this is likely due to zooming out introduce padding rather than true equviariance to scale changes. We also observe equivariance to appearance transformations for ResNet18 such as posterize that are also absent from ViT.

\subsection{Classes consistently hurt by \texttt{RandomResizedCrop}}
\label{sec:apphurt}
\begin{table}[h!]
	\centering
    \begin{tabular}{l|c|c}
    Class & \% of comparisons where hurt & Loss in Top-1 accuracy (\%) \\
    \hline
cassette player & 100.0 & 22.00 \\
maillot & 100.0 & 21.20\\
palace & 100.0 & 4.40\\
academic gown & 92.0 & 13.57\\
missile & 88.0 & 11.82\\
mashed potato & 88.0 & 9.09\\
digital watch & 84.0 & 7.52\\
barn spider & 80.0 & 7.70\\
Indian elephant & 80.0 & 7.20\\
miniskirt & 80.0 & 8.00\\
pier & 80.0 & 5.30\\
wool & 80.0 & 6.80\\
ear & 80.0 & 10.60\\
brain coral & 72.0 & 3.89\\
crate & 72.0 & 4.33\\
fountain pen & 72.0 & 5.22\\
space bar & 72.0 & 6.33\\
\end{tabular}
\caption{Validation classes hurt by \texttt{RandomResizedCrop}}
\label{tab:hurtclasses}
\end{table}

We compare the per-class top-1 accuracy when using \texttt{RandomResizedCrop} augmentation training versus FixedSized CenterCrop augmentation (which is the no augmentation). On average, over 25 pairwise comparisons of 5 runs with both augmentations, $12.3\% \pm 0.21\%$ of classes are hurt by the use of \texttt{RandomResizedCrop}. We list in Table \ref{tab:hurtclasses} the classes that are hurt in more than $70\%$ of the 25 comparisons, and the average amount of decrease in Top-1 accuracy when incurred. We do not see a pattern in these classes when reading their labels. We confirm the lack of pattern by computing the similarities of the classes listed in Table \ref{tab:hurtclasses} using the most specific common ancestor in the Wordnet \citep{wordnet} tree. The similarity of the classes that are consistently hurt (17 classes) is $0.42 \pm 0.013$ ($ 0.49 \pm 0.019$ if we include a class similarity with itself), while the similarity between the classes that are consistently hurt and the ones that are consistently helped ($> 70 \%$ of comparisons) is $0.406 \pm 0.0014$, and between classes consistently hurt and everything else (neither helped or hurt) is $0.403 \pm 0.002$.

\section{Similarity Search}
\label{sec:appSimSearch}

\paragraph{Experimental details} We use the same trained models using standard augmentations as we did for equivariance. For the no augmentation ResNet18, we use the training procedure outlined in Appendix \ref{sec:trainingdetails}. We sample $1k$ pairs from each class and compute SimChange$_{T_\theta}$ for each pair. We pool transformations from subpolicies discovered by AutoAugment on ImageNet, SVHN, and additional rescaling for zooming in and out. We extend the implementation of AutoAugment provided in the DeepVoltaire library \url{https://github.com/DeepVoltaire/AutoAugment }. Note we disregard the learned probabilities from AutoAugment and instead apply each transformation independently for our similarity search analysis. For subpolicies, we apply each transformation in sequence. Transformations include `equalize', `solarize', `shearX', `invert', `translateY', `shearY', `color', `rescale', `autocontrast', `rotate', `posterize', `contrast', `sharpness', `translateX' with varying magnitudes. We illustrate the effect of transformations in Fig. \ref{fig:transforms}.

\paragraph{No transformation consistently increases pair similarity} In Fig. \ref{fig:appSimChange} we show the distribution of similarity change of each transformations over classes. While for some outlier classes, some transformations increase similarity among pairs, distributions are below or near 0 for all transformations. The top single transformation across all classes is posterize, which increases similarity by $0.02 \pm 0.02 (SEM)$, implying no statistically significant increase. In contrast, we find on average 6.5 out of the top 10 transformations \textit{per class} increase similarity by $5.4\% \pm 0.04\%(SEM)$.

\begin{figure}[h!]
    \centering
    \includegraphics[width=\textwidth]{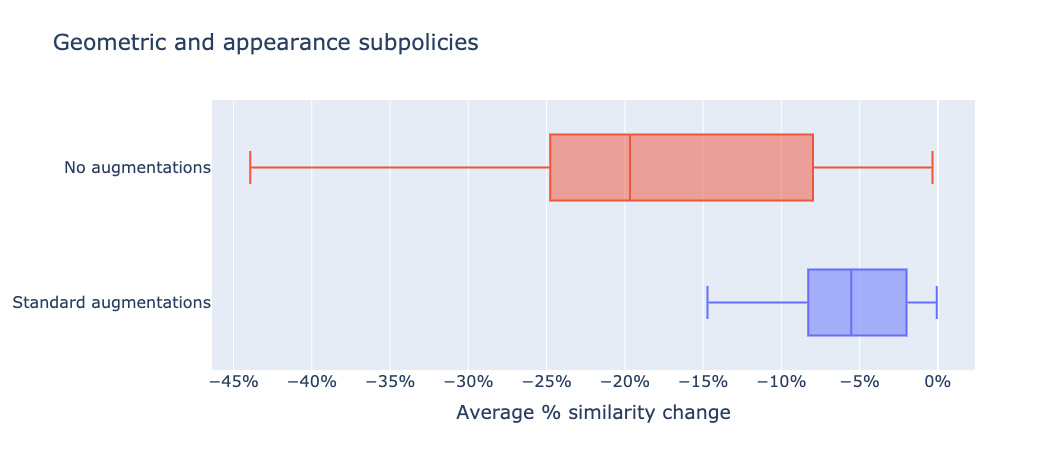}
    \caption{Similarity Search average change across classes for the best subpolicies discovered by AutoAugment on ImageNet. We see no subpolicy consistently increases similarity among pairs.}
    \label{fig:appSimSearchSubpolicies}
\end{figure}

\begin{figure}[h!]
    \centering
    \includegraphics[width=\textwidth]{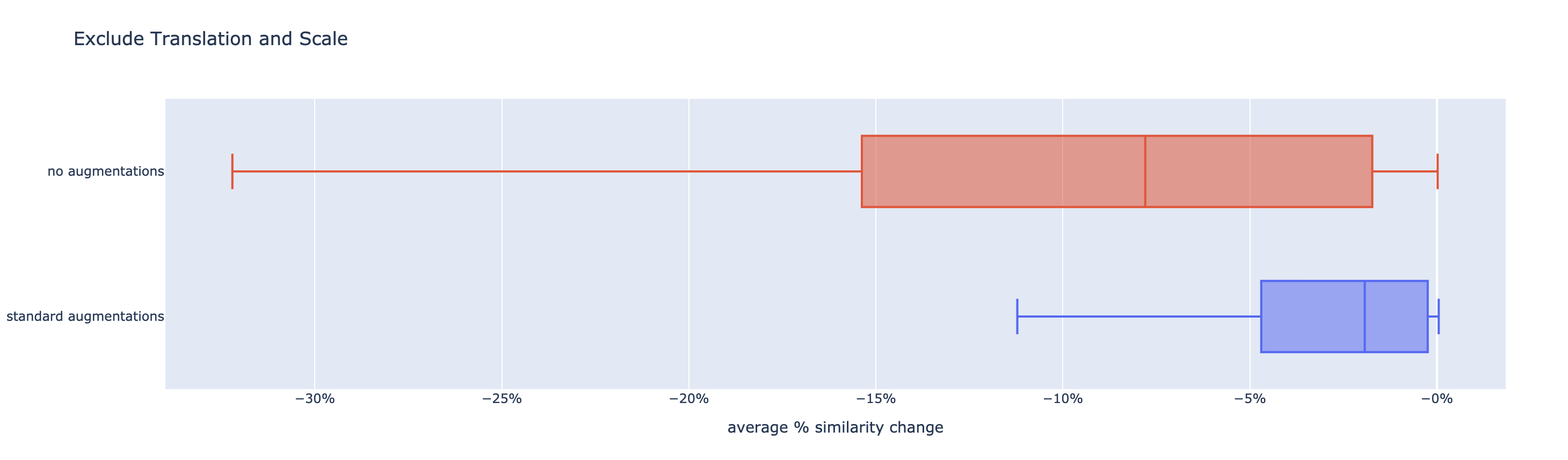}
    \caption{Similarity Search excluding translation and scale which are used during standard augmentation.}
    \label{fig:simSearchNotStandard}
\end{figure}

\paragraph{Geometric transformations} We also examine the distributions for transformations not in the standard augmentations by excluding all translations and rescales. In Fig. \ref{fig:simSearchNotStandard} we see even if we exclude translation and scale, standard data augmentation drastically decrease the variation of similarity changes even for other transformations not used during training.

\subsection{Measuring local variation in foregrounds}
\label{sec:appLocalVariation}
To supplement our analysis of global transformations, we also directly measure local variation in ImageNet using foregrounds extracted from U2Net \citep{Qin_2020} trained on DUTS  \citep{wang2017}. We measure the center coordinates and area of bounding boxes around the foreground object relative to the image frame using a threshold of $0.01$ to determine the bounding box. We measure foreground variation over all training images in ImageNet. We observe there is more variation in scale, which ranges from $31\%-74\%$ of the image, compared to translation which is centered ($50\% \pm 5\%$ (SEM)).

\begin{figure}[h!]
        \centering
    \begin{subfigure}[b]{\textwidth}
        \centering
        \includegraphics[width=\textwidth]{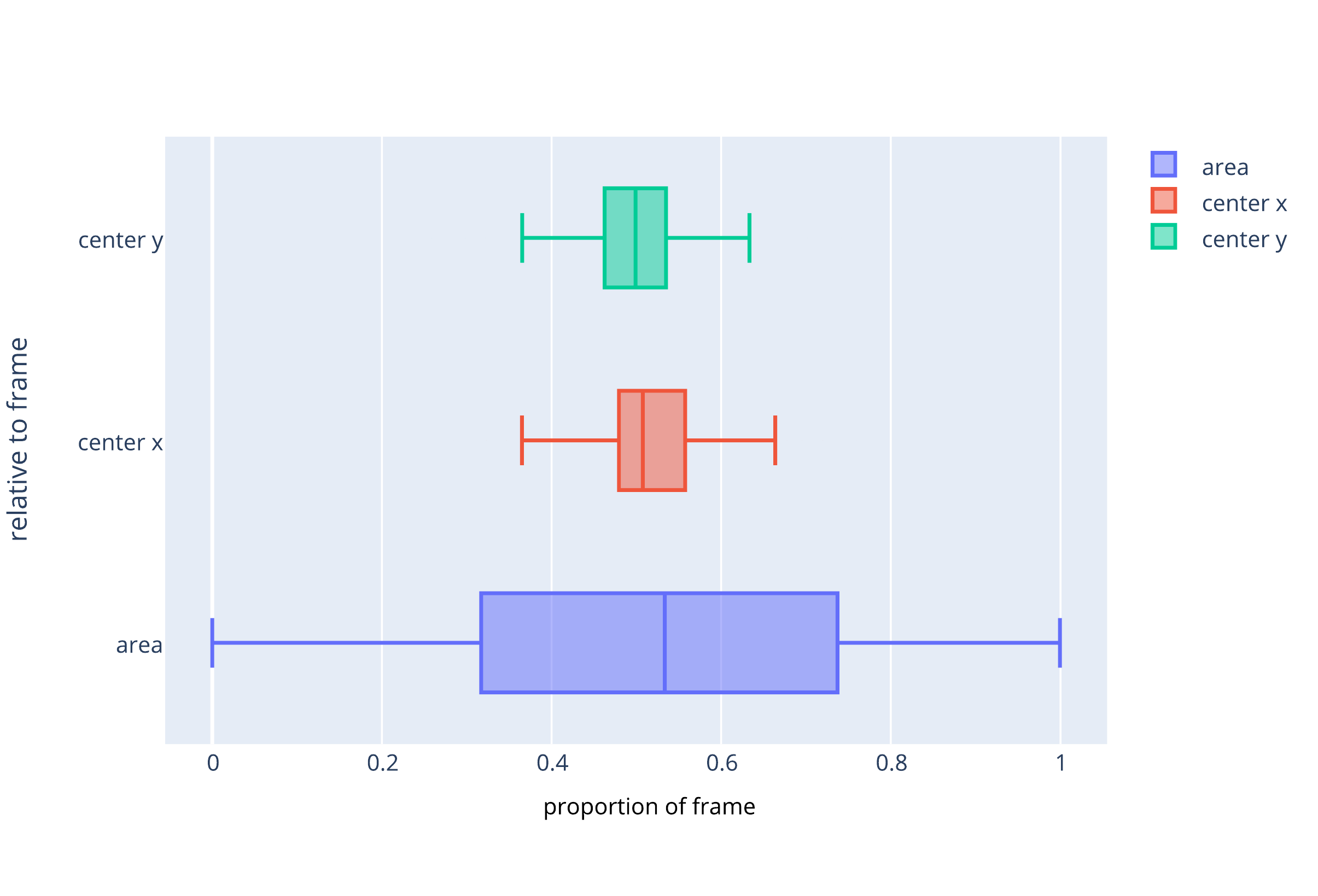}
    \end{subfigure}
    \caption{Local variation in foregrounds.}
    \label{fig:foregroundBoxes}
\end{figure}
\subsection{Textiles Weighted Boost}
\label{app:simSearchWeightedBoost}

To account for both the size of the similarity increases and the proportion of images increased, we rank transformations by their weighted boost, defined as the average percent boost * proportion of image pairs boosted. We examine the top 10 transformations by weighted boost and find rotate with magnitudes ranging from 3-9 is the top transformation for all top 10 for the ResNet18 trained with standard augmentation. We find for rotation the corresponding classes are velvet, handkerchief, envelop, and wool with velvet appearing 4 times among the top 10. For subpolicies, we find both the ResNet18 trained with or without standard augmentations, rotation and a color transformation with varying magnitudes is the top 10 transformation also corresponding to velvet, wool, handkerchief, and envelope with jigsaw puzzle as an additional class. 

\subsection{Wordnet Similarity Search}
\label{sec:appSimSearchWordnet}

To study whether similar class have similar factors of variation, we measured class similarity using the WordNet hierarchy. We compute similarity using several methods provided in the NLTK library \url{https://www.nltk.org/howto/wordnet.html} including Wu-Palmer score, Leacock-Chodorow Similarity, and path similarity---all of which compute similarity by comparing the lowest common ancestor in the WordNet tree. To compute similarity of transformations, we compared the Spearman rank correlation of the top transformation in each class by average percent similarity change as well as proportion of image pairs boosted. We found no significant difference between the two. We compare class WordNet similarity against transformation ranks for all $1k$ ImageNet classes.

\section{Additional experiment: How does augmenting validation images affect accuracy?}

Standard data augmentation methods are a proxy to implement geometric transformations such as translation and scale, do they actually bring invariance to these transformations? If so, the performance of a model trained with data augmentation should be equal even when validation images are augmented. To answer this question, similar to \citet{Engstrom19}, we augment the images during evaluation to test for scale invariance. Indeed, robustness to augmentations is used a generalization metric in \citet{Aithal2021}.\\
\\
Specifically, we augment the validation images with the regular validation pre-processing\texttt{ FixedSixeCenterCrop}, and then augment the images by taking a \texttt{RandomSizeCenterCrop} (disabling aspect ratio change). This scales the object in the crop. Using \texttt{RandomSizeCenterCrop}, we vary $s_{-}$, that specifies a lower bound of the uniform distribution $s \sim U(s_{-},1)$. This effectively varies the maximal increase in size potentially applied, which is proportional to $1/s_{-}$. We compare the models trained with \texttt{\texttt{RandomResizedCrop}} and \texttt{FixedSizeCenterCrop} (no augmentation). 
\begin{figure}[h!]
	\centering
	\includegraphics[width=0.6\textwidth]{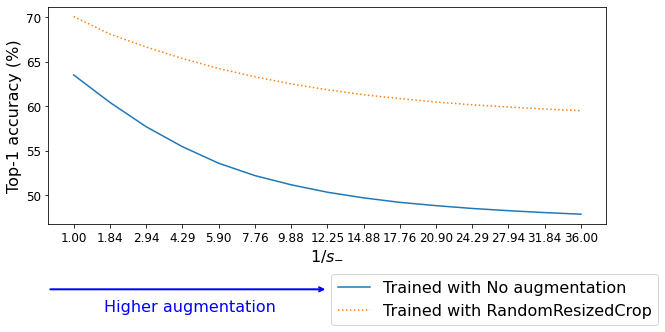}
    \caption{Augmenting the validation set.}
    \label{fig:invarianceplot}
\end{figure}

Specifically, we select $15$ values of increase in size per-axis $v$ uniformly between $1$ and $6$, and set $s_{-}=1/v^2$. Thus we augment the images from no augmentation ($s=1$) to scaling the image by a factor potentially as large as $36$ when $v=6$. Note that for the value $v=3.5$, $s_{-}=1/v^2 \approx 0.08$, matching the value of \texttt{RandomResizedCrop} lower bound on the range of $s$. To match the procedure done at evaluation, we use \texttt{FixedSizeCenterCrop}, and then augment the crop. For each model, we evaluate the $5$ training seeds of the best hyper-parameters setting, running each augmentation experiment for $5$ different test seeds. We compute performance averaged over the test seeds and then report the mean of the $5$ training seeds $\pm$ standard error of the mean.\\
\\
Fig. \ref{fig:invarianceplot} shows that the accuracy on the validation set decreases as we increase the magnitude of augmentations on validation images. The stronger decrease is for the no augmentation model, while the model trained with \texttt{\texttt{RandomResizedCrop}} is more robust. However, while the latter has been trained with augmentations up to $1/s_{-}\approx12.5$, its performance already decreases for the smallest values\footnote{Note that we apply \texttt{FixedSizeCenterCrop}, which resizes the image to $256$ on the shorter dimension, before \texttt{RandomSizeCenterCrop}. Thus, compared to what is done at training, there is an additional scaling of $256^2/224^2\approx1.3$ for the same value $s$.}. This suggests it might only present \emph{partial} invariance.

\section{Additional experiment: Augerino \citep{benton2020learning} on ImageNet} 
\label{sec:augerino}
We have shown that standard data augmentation methods rely on a precise combination of transformations and parameters, and needs to be hand-tuned. To overcome these issues, and potentially discover relevant factors of variation of the data, recent methods have been proposed to automatically discover symmetries that are present in a dataset \citep{benton2020learning,zhou_meta-learning_2020,dehmamy2021lie}. We assess the potential of a state-of-the-art model of this type to tackle ImageNet, that is, the Augerino model \citep{benton2020learning}.

\subsection{The Augerino model and our modifications}
\label{sec:augerinodetails}
\paragraph{Augerino method} Augerino is a method for automatic discovery of relevant equivariances and invariances from training data only, given a downstream task. Given a neural network $f_w$ parametrized by $w$, Augerino creates $\hat{f}$ a model approximately invariant to a group of transformations $G$ by averaging the outputs over a distribution $\mu_{\theta}$ over $g \in G$:
\begin{equation}
    \hat{f}_w(x) = \mathbb{E}_{g \sim \mu_{\theta}}[f_w(gx)].
\end{equation}
Augerino considers the group of affine transformations in 2D, Aff(2), composed of 6 generators corresponding to translation in $x$, translation in $y$, rotation, uniform scaling, stretching and shearing.\footnote{While in the original paper \citet{benton2020learning} mentions scale in $x$ and scale in $y$ and shearing, the generators of Aff(2), employed in the paper, in fact correspond to uniform scaling, stretching and shearing.}.
\\
Instead of directly using a distribution over transformations in image space $g$, the distribution is parametrized over \emph{the Lie algebra of $G$}. For insights on Lie Groups and Lie algebras, we refer the interested reader to \citet{hall2015lie}. Thus, $\theta$ specifies the bounds of a uniform distribution in the Lie algebra. It is $6$-dimensional, each $\theta_i$ specifies the bounds of the distribution over the subgroup $G_i$: $U(-\theta_i/2,\theta_i/2)$. When the value sampled in the Lie algebra is $0$, this corresponds to the transformation $g$ being the identity. The smaller $\theta_i$, the smaller the range transformations in $G_i$ are used, and a dirac distribution on $0$ always returns the identity transformation, i.e. no use of the transformation $G_i$.\\
\\
The value of $\theta$ is learned along the parameters of the network $w$, specifying which transformation is relevant for the task at hand. In the case of classification, the cross-entropy loss is linear and thus expectation can be taken out of the loss. 
\begin{equation}
    l(\hat{f}_w(x)) = l(\mathbb{E}_{g \sim \mu_{\theta}}[f_w(gx)]) = \mathbb{E}_{g \sim \mu_{\theta}}[l(f_w(gx)].
\end{equation}
 Furthermore, Augerino employs a negative $L2$-regularization on $\theta$, parametrized by $\lambda$, to encourage wider distributions. The resulting training objective is:
\begin{equation}
   \min_{\theta,w}~\mathbb{E}_{g \sim \mu_{\theta}}[l(f_w(gx))] - \lambda ||\theta||_2
   \label{eq:augerinoto2}
\end{equation}
where a larger $\lambda$ pushes for wider uniform distribution. Everytime an image is fed to the model, Augerino (1) draws a sample from the uniform distribution on the Lie algebra of $G$ (2) computes the transformation matrix in image space through the exponential map (3) augments the image with the corresponding transformation (4) feeds the augmented image to the neural network to perform classification. See \citet{benton2020learning} for more details on the model.

\paragraph{Regularizing by transformations} We noticed that the original code library of \citet{benton2020learning} disables regularization if any of the $6$ coordinates in $\theta$ (one per each of the transformation in Aff(2)) has reached a certain value. From our initial experiments, we understand this is to prevent underflow errors. However, we find it to be too strict if we want to learn multiple transformations at the same time, and thus we shutdown regularization on each $\theta_i, i=1,\ldots,6$ when a specific value for that coordinate $i$ is reached.\footnote{If the value goes below in the next iteration, regularization is enabled again.} This also allows us to control separately each transformation regularization. While \citet{benton2020learning} has found that their model was insensitive to the regularization strength, we find it to be a key parameter when applied to ImageNet. This is a first difficulty we face when trying to tackle a real dataset with an equivariant model. While the original model considers always all $6$ possible transformations, we also modified the original model to consider any transformation separately and any of their combinations.

\subsection{Augerino on translation discovery}
We study if Augerino discovers translation as a relevant transformation to improve performance on ImageNet. We shutdown regularization if the bound of the distributions (separately for each x and y-axis) has reached a value corresponding to $-50\%,50\%$ translation in image space. We cross-validate between multiple seeds and bounds regularization parameters. As Augerino is employed at validation, we perform testing with $5$ different seeds. We also show in Table \ref{tab:augerinotxtydisabled} the results of Augerino disabled at evaluation.
\begin{table}[h!]
    \RawFloats
	\begin{minipage}{0.4\textwidth}
	\centering
    \begin{tabular}{l|c}
    $\lambda$ & {Top-1 $\pm$ SEM}\\
    \hline
    0.01 & $63.53\pm 0.0$\\
    0.1 & $63.78\pm 0.1$\\
    0.2 & $64.3\pm 0.1$\\
    0.4 & $\mathbf{67.22\pm 0.0}$\\
    0.6 & $67.05\pm 0.0$\\
    0.8 & $67.01\pm 0.0$\\
    1.0 & $66.98\pm 0.0$\\
    \end{tabular}
    \captionof{table}{ImageNet validation set Top-1 accuracy $\pm$ standard error of the mean (SEM) over training seeds, for different $\lambda$.}
    \label{tab:augerinotxty}
    \end{minipage}
	\begin{minipage}{0.6\textwidth}
	\centering
	\includegraphics[width=\textwidth]{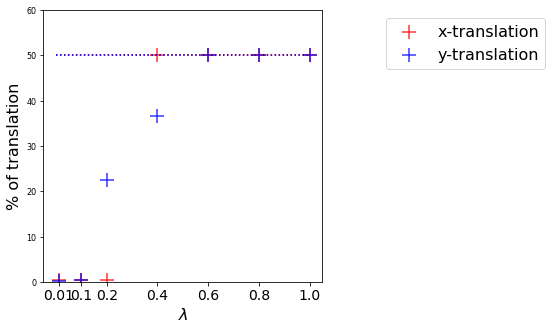}
    \captionof{figure}{Learned translations versus $\lambda$ values. Every marker correspond to a learned value. Dashed lines correspond to $50\%$ i.e. when the regularization is disabled.}
    \label{fig:augerinotxty}
    \end{minipage}
\end{table}
Table \ref{tab:augerinotxty} shows that for different values of the regularization parameter $\lambda$ (see Equation \ref{eq:augerinoto2}), different bounds are learned by the model. In Fig. \ref{fig:augerinotxty} we compare the learned bounds of the distribution. The bounds saturate at a value corresponding in image space to $50\%$, i.e. when the regularization is disabled (shown in dashed lines). This shows that, contrarily to the experiments in \citet{benton2020learning}, the regularization term in Equation \ref{eq:augerinoto2} strongly impacts the learned bounds. Best performance is achieved for $\lambda=0.4$, with learned bounds that correspond to sampling a translation in $\approx[-50\%,50\%]$ on the x-axis, and $\approx[-36.7\%,36.7\%]$ on the y-axis. Augerino discovers translation as a relevant augmentation, and learn values that improve over the \texttt{FixedSizeCenterCrop} (no augmentation) method.\footnote{We use Augerino on top of \texttt{FixedSizeCenterCrop} pre-processing. For comparison, a model trained with T.(30\%) with \texttt{FixedSizedCenterCrop} pre-processing achieves $67.08\pm 0.1$ Top-1 accuracy.} 

Table \ref{tab:augerinotxtydisabled} shows the performance of the model trained with Augerino for translation discovery, when Augerino is not employed during evaluation time. We do note they are slightly higher (by up to $\approx 1\%$ for $\lambda \ge 0.6$) to the ones reported in Table \ref{fig:augerinotxty}.

\begin{table}[h!]
	\centering
    \begin{tabular}{l|c}
    $\lambda$ & {Acc@1 $\pm$ SEM}\\
    \hline
    0.01 & $63.31\pm 0.1$\\
    0.1 & $63.49\pm 0.1$\\
    0.2 & $64.08\pm 0.1$\\
    0.4 & $67.68\pm 0.1$\\
    0.6 & $\mathbf{67.98\pm 0.0}$\\
    0.8 & $\mathbf{67.95\pm 0.1}$\\
    1.0 & $\mathbf{67.99\pm 0.0}$\\
    \end{tabular}
    \caption{ImageNet validation set Top-1 accuracy $\pm$ standard error of the mean (SEM) for Augerino using translation in x and y axes, for different $\lambda$. These results are when Augerino is disabled at evaluation time (no augmentation of the validation images).}
    \label{tab:augerinotxtydisabled}
\end{table}

\subsection{Augerino on translation and scale} 
\label{sec:augtxtyscale}
When we use Augerino on the scale-translation group, we want to control the parameters so that we can use specific shutdown values corresponding to $50\%$ translation and $2000\%$ scaling, and apply translation before scaling as is done in the \texttt{T.(30\%) + RandomSizeCenterCrop} method. Thus when we use the Augerino model to learn parameters for the scale-translation group, we performed a few modifications to the original code. First, we explicitly compute translation before scaling. Second, Augerino's code uses the \emph{affine\_grid} and \emph{grid\_sample} methods in Pytorch, where the former performs an inverse warping. That is, for a given scale $s$, the inverse scaling is performed. This does not impact the translation which is performed in both direction (negative and positive translation). For scale, we shutdown regularization if the value $\theta_s$ corresponds to $s=1/2000=0.05\%$, as the inverse scaling will be performed. Third, we want to sample scales corresponding to an increase in size (zooming-in) in order to mimic the effect of \texttt{RandomResizedCrop} which selects only a subset of the image. Hence we take $U(-\theta_s/2, 0)$. A sampled value which is negative corresponds to a positive scaling by Augerino given the inverse wrapping. We shutdown the regularization parametrized by $\lambda$ when translation has reached $50\%$ of the width / height, and a scaling of $2000\%$ (the object appears $20$ times larger). \\
\\
Table \ref{tab:augtxtyscale} shows the performance of Augerino when trained on ImageNet with the possibility to learn about translation and scale in conjunction. Best performance is achieved with $\lambda=0.2$, and we see in Fig. \ref{fig:augtxtyscale} that the model has learned to augment with x-translations only (the red marker being close to $\approx 40\%$ and the green and blue marker being close to $0$). Interestingly, when Augerino is disabled at evaluation time, Table \ref{tab:augerinotxtydisabled} shows that the Top-1 accuracy is much higher for large values of regularization compared to Table \ref{tab:augtxtyscale}, with the best performing $\lambda=0.6$. This means that with a more aggressive augmentation, the use of Augerino hurts during inference and only slightly helps during training compared to the no augmentation case (see Table \ref{tab:augmentfull} for \texttt{FixedSizeCenterCrop}). Still, as in the experiment for translation only, we note that the value of $\lambda$ greatly impacts the results, and that the gain in performance compared to no augmentation is quite small. More importantly, the performance is smaller than when using translation only (see Table \ref{tab:augerinotxty}), which the model could have fallback on if scale was not a relevant transformation. For comparison, a model trained with \texttt{T.(30\%) + RandomSizeCenterCrop} with \texttt{FixedSizedCenterCrop} pre-processing achieves $67.33\pm 0.0$ Top-1 accuracy. We conclude that while Augerino is a promising model for automatic discovery of relevant symmetries in data, it remains a challenge to apply such methods on a real, large-scale dataset such as ImageNet.

\begin{table}[h!]
\RawFloats
	\begin{minipage}{0.4\textwidth}
	\centering
    \begin{tabular}{l|c}
    $\lambda$ & {Acc@1 $\pm$ SEM}\\
    \hline
    0.01 & $63.51\pm 0.1$\\
    0.1 & $63.73\pm 0.1$\\
    0.2 & $\mathbf{63.90\pm 0.1}$\\
    0.4 & $63.34\pm 0.0$\\
    0.6 & $62.83\pm 0.0$\\
    0.8 & $61.0\pm 0.1$\\
    1.0 & $57.71\pm 0.0$\\
    \end{tabular}
    \captionof{table}{ImageNet validation set Top-1 accuracy $\pm$ standard error of the mean (SEM) for different $\lambda$.}
    \label{tab:augtxtyscale}
    \end{minipage}
	\begin{minipage}{0.6\textwidth}
	\centering
	\includegraphics[width=\textwidth]{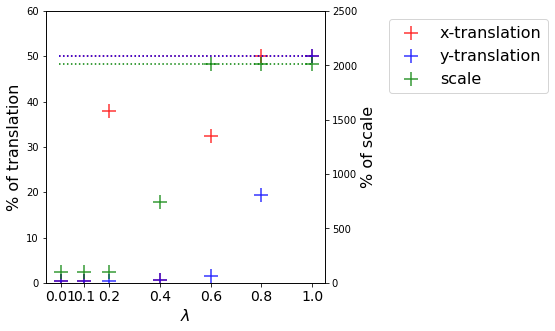}
    \captionof{figure}{Learned translations versus $\lambda$ values.}
    \label{fig:augtxtyscale}
    \end{minipage}
\end{table}

\begin{table}[h!]
	\centering
    \begin{tabular}{l|c}
    $\lambda$ & {Acc@1 $\pm$ SEM}\\
    \hline
    0.01 & $63.11\pm 0.1$\\
    0.1 & $63.29\pm 0.1$\\
    0.2 & $63.7\pm 0.1$\\
    0.4 & $63.40\pm 0.1$\\
    0.6 & $\mathbf{64.16\pm 0.1}$\\
    0.8 & $63.41\pm 0.1$\\
    1.0 & $62.27\pm 0.1$\\
    \end{tabular}
    \captionof{table}{ImageNet validation set Top-1 accuracy $\pm$ standard error of the mean (SEM) for different $\lambda$, disabling Augerino at evaluation.}
    \label{tab:augtxtyscaledisabled}
\end{table}

\end{appendices}

\end{document}